\documentclass[preprint,12pt]{elsarticle}

\usepackage{graphicx}
\usepackage{amssymb}
\usepackage{float}
\usepackage{lineno,hyperref}
\usepackage{amsmath,amsfonts,amsthm,amssymb,graphicx,subfigure,mathtools,tikz,hyperref,bm,blindtext}
\usepackage{diagbox}
\usepackage{geometry}
\usepackage[acronym]{glossaries}
\usepackage{multirow}
\usepackage{algorithm}
\usepackage{algpseudocode}
\usepackage{bibentry}
 
\usepackage{lineno}

\journal{Journal Name}

\begin{document}

\begin{frontmatter}


\title{Multi-fidelity Bayesian Neural Networks: Algorithms and Applications}



\author[brown]{Xuhui Meng}
\author[pitt]{Hessam Babaee}
\author[brown,PNNL]{George Em Karniadakis \fnref{2}}

\fntext[2]{Corresponding author: george\_karniadakis@brown.edu (George Em Karniadakis).}
\address[brown]{Division of Applied Mathematics, Brown University, Providence, RI 02906, USA}
\address[pitt]{Mechanical Engineering and Materials Science, University of Pittsburgh, Pittsburgh, PA 15260, USA}
\address[PNNL]{Pacific Northwest National Laboratory, Richland, WA 99354, USA}

\begin{abstract}
We propose a new class of Bayesian neural networks (BNNs) that can be trained using noisy data of variable fidelity, and we apply them to learn function approximations as well as to solve inverse problems based on partial differential equations (PDEs). These multi-fidelity BNNs consist of three neural networks: The first is a fully connected neural network, which is trained following the maximum a posteriori  probability (MAP) method to fit the low-fidelity data; the second is a Bayesian neural network employed to capture the cross-correlation with uncertainty quantification between the low- and high-fidelity data; and the last one is the physics-informed neural network, which encodes the physical laws described by PDEs. For the training of the last two neural networks, we first employ the mean-field variational inference (VI) to maximize the evidence lower bound (ELBO) to obtain informative prior distributions for the hyperparameters in the BNNs, and subsequently we use  the Hamiltonian Monte Carlo method to estimate accurately the posterior distributions for the corresponding hyperparameters. We demonstrate the accuracy of the present method using  synthetic data as well as real measurements. Specifically, we first approximate a one- and four-dimensional function,  and then infer the reaction rates in one- and two-dimensional diffusion-reaction systems. Moreover,  we infer the sea surface temperature (SST) in the Massachusetts and Cape Cod Bays using satellite images and in-situ measurements. Taken together, our results demonstrate that the present method can capture both linear and nonlinear correlation between the low- and high-fideilty data adaptively,
identify unknown parameters in  PDEs, and quantify uncertainties in predictions, given a few scattered noisy high-fidelity data. Finally, we demonstrate that we can effectively and efficiently reduce the uncertainties and hence enhance the prediction accuracy with an active learning approach,  using as examples a specific one-dimensional function approximation and an inverse PDE problem.

\end{abstract}

\begin{keyword}
auto-regressive scheme \sep nonlinear correlation \sep physics-informed neural networks \sep Hamiltonian Monte Carlo \sep uncertainty quantification  \sep active learning  \sep satellite data


\end{keyword}

\end{frontmatter}



\section{Introduction}
 In many applications in science and engineering the accuracy of predictions by computational models may vary, depending on the the available multi-fidelity data \cite{PWG18,fernandez2016review}. Typically, increased fidelity of the data sources is associated with higher acquisition cost. Expensive computations may include tasks such as uncertainty quantification, optimization, inverse problems and sensitivity analysis, in which a computational model must be evaluated in an outer loop \cite{FSK07,Babaee:2013ab,Perdikaris:2015aa,Perdikaris20151107}.  For these systems, various data sources for obtaining predictions may be available such as fine-grid/coarse-grid  numerical models or reduced order models.  In other applications, the data  may come both from both a numerical model and experimental measurements. In the latter case, the data may be obtained from different types of instruments;  for example, satellite images and in-situ measurements can provide data sources for inferring sea surface temperature (SST).  In a traditional \emph{single data source} approach, a decision has to be made between using low-fidelity predictions with dense samples, which  can yield a high-resolution  input-output map versus sparser high-fidelity predictions, since they are costly to acquire.  
 
Multi-fidelity techniques provide a \emph{data multi-source} paradigm, in which \emph{all}  sources of data with variable fidelity are utilized to enhance predictions. Auto-regressive Gaussian processes regression (GPR) is one of the most commonly used Bayesian techniques for multi-fidelity modeling \cite{KO00,BPCK16}, yielding predictions of the mean values as well as estimates of uncertainty.  Predicting uncertainty is critical in many applications, especially for cost-effective allocation of  limited data acquisition resources.  The fact that GPR multi-fidelity models provide uncertainty quantification is one of the main reasons of the popularity of these techniques; see, for example \cite{BBDCK20}, where the GPR multi-fidelity model for  SST is built by combining SST satellite measurements (low-fidelity) with in-situ measurements (high-fidelity). The GPR uncertainty maps obtaoned could then be utilized  for guiding future in-situ measurements using an appropriate acquisition function \cite{MFBC09}.

One of the challenges of GPR multi-fidelity models is that  the computational complexity of training multi-fidelity GPR scales as $\mathcal{O}(N^3)$, where $N$ is the total number of high-fidelity and low-fidelity training points. As a result, the utility of GPR is largely limited to problems with low dimensional input space corresponding to a small number of training points. Moreover,  GPR multi-fidelity models seek  to exploit linear correlations between various levels of fidelity. Although an extension of the GPR multi-fidelity model to extract nonlinear correlations has been developed in~ \cite{PRDLK17}, this comes at a cost of adding one more dimension to the input space, and, therefore, it requires larger datasets for training.  As a remedy to these limitations,   recently, a multi-fidelity scheme based on a composite neural network (NN) was presented \cite{meng2020composite}, in which both linear and nonlinear correlations between low-fidelity and high-fidelity data sets are exploited. The minibatch training of the NN enables the application of this technique to large training datasets in a scalable manner.  Furthermore, the composite neural network (NN) developed  in \cite{meng2020composite} has also been successfully applied to solve inverse PDE problems based on the idea of physics-informed neural networks \cite{raissi2019physics}. However, the existing multi-fidelity model using the composite NN does not predict uncertainty, which is a byproduct of  Bayesian multi-fidelity models. This is the issue we address in the current work.

Specifically, in this paper we present a multi-fidelity scheme based on Bayesian neural networks. The novelty of the framework  
we propose is three-fold: (i) it obtains  good uncertainty estimates for the quantity of interest, e.g., high-fidelity predictions, unknown parameters in PDEs, etc.; (ii) it exploits both linear and nonlinear correlations between low-fidelity and high-fidelity data adaptively; and (iii) it is scalable with respect to the size of the training datasets. The paper is organized as follows:  we first present the details of the proposed multi-fidelity approach in Sec. \ref{sec:method}, while the results including function approximations, inverse PDE problems, and active learning are included in Sec. \ref{sec:results};  a summary of the present study is given in Sec. \ref{sec:summary}.



\section{Methodology}
\label{sec:method}
\subsection{Multi-fidelity Bayesian neural networks}
\label{sec:mbnn}
\begin{figure}[H]
\centering
\includegraphics[width=0.9\textwidth]{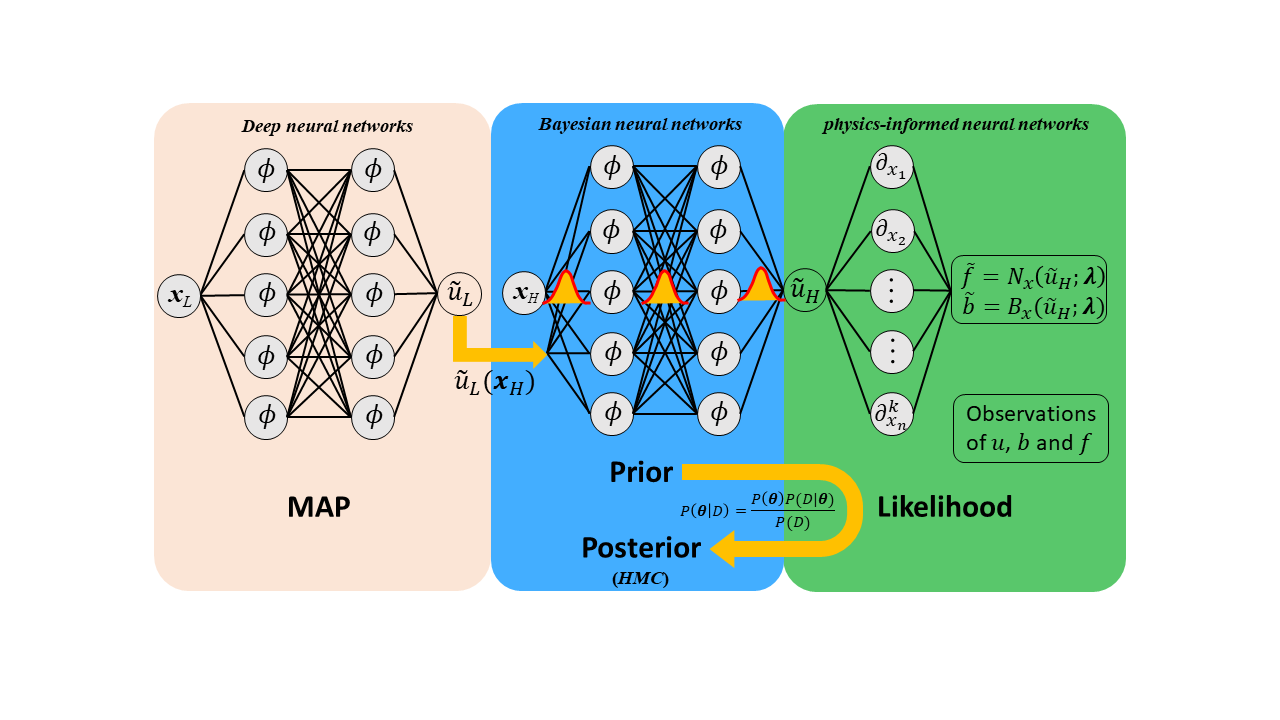}
\caption{
Schematic of multi-fidelity Bayesian neural networks. The orange panel on the left represents the deep neural network (DNN) for low-fidelity data, the blue panel represents the Bayesian neural network (BNN) for the cross-correlation, and the green panel is the physics-informed part. $P(\bm{\theta})$ is the prior distribution for hyperparameters in BNNs and the unknowns in PDEs, $P(\mathcal{D}|\bm{\theta})$ represents the likelihood of observations (e.g., $u, b, f$), $P(\bm{\theta}|\mathcal{D})$ is the posterior, and $P(\mathcal{D})$ denotes the marginal likelihood. $\bm{x}_L/\bm{x}_H$: locations of the low-/high-fidelity training data, $\tilde{u}_L$/$\tilde{u}_H$: low- and high-fidelity predictions, $\tilde{u}_L(\bm{x}_H)$: low-fidelity predictions at $\bm{x}_H$, $\phi$: activation function, which is the hyperbolic tangent function in this study. MAP: maximum a posteriori probability, HMC: Hamiltonian Monte Carlo. 
}
\label{fig:bnn}
\end{figure}

As shown in Fig. \ref{fig:bnn}, the multi-fidelity Bayesian neural network is composed of three different neural networks: the first is a deep neural network (DNN) to approximate the low-fidelity data, while the second one is a Bayesian neural network (BNN) for learning the correlation (with uncertainty quantification) between the low- and high-fidelity data. The last one is the physics-informed neural network (PINN), which is employed to encode the physical laws represented by partial differential equations (PDEs) \cite{raissi2019physics,yang2020b}.

Considering that the low-fidelity data are generally plentiful, the computational cost is prohibitive if a BNN is used here. To achieve high computational efficiency, we therefore employ the maximum a posteriori probability estimate (MAP) to train the first neural network (NN), which is the same as in \cite{meng2020composite}. Each hyperparameter in the first neural network now is a point estimate rather than a distribution. The loss function for the MAP is expressed as
\begin{align}\label{eq:loss_l}
    \mathcal{L}_L = \frac{1}{N_L}|\bm{u}_L - \tilde{\bm{u}}_L|^2 + \alpha |\bm{w}_L|^2,
\end{align}
where $N_L$ is the number of low-fidelity training data, $\bm{u}_L$ and $\tilde{\bm{u}}_L$ are low-fidelity training data and predictions, respectively, the second term is the $\mathcal{L}_2$ regularization with $\bm{w}_L$ denoting the weights in the first DNN, with $\alpha$ the weight of the regularization. Note that the $\mathcal{L}_2$ regularization here is capable of reducing the overfitting caused by the measurement noise as well as the DNN itself.

We then  employ the Bayesian neural network \cite{neal2012bayesian} (Fig. \ref{fig:bnn}), which takes $(\bm{x}_H, \tilde{\bm{u}}(\bm{x}_H))$ as inputs to quantify the uncertainties in predicting the correlations between the low- and high-fidelity data. In particular, the Hamiltonian Monte Carlo (HMC) method \cite{neal2011mcmc}, which is frequently used as ground truth for posterior estimation, is adopted here to estimate the posterior distributions in BNN \cite{yao2019quality}. In general, the estimation of posterior distributions for hyperparameters in the BNN is computationally prohibitive for problems with big data  using HMC \cite{yang2020b}. However,  the high-fidelity data are generally scarce due to high acquisition cost associated with high-fidelity data. Therefore, we can exploit the BNN to learn the correlation between the low- and high-fidelity data with uncertainty quantification given a small amount of high-fidelity observations at low cost. Details for the training of BNNs for function approximation as well as inverse PDE problems will be elaborated below. Note that all variables without subscripts in this study denote the measurements/predictions at the high-fidelity level (as well as variables with subscript ``H").

We assume that for function approximation, the high-fidelity datasets are noisy scattered measurements on $u$, for example from sensors, i.e., $\mathcal{D} = \mathcal{D}_u$. In addition, we assume that the measurements are independent Gaussian distributions, i.e.,
\begin{equation}
    \begin{aligned}
    {u}^{(i)} = \bar{u}(\boldsymbol{x}_{u}^{(i)}) + \epsilon_u^{(i)}, \quad i = 1,2...N_u, 
    \end{aligned}
\end{equation}
where $\bar{u}(\boldsymbol{x}_{u}^{(i)})$ and $\epsilon_u^{(i)}$ are the exact value and the independent Gaussian noise with zero mean. Furthermore,  the noise scale of each sensor is assumed to be known, i.e., the standard deviations of $\epsilon_u^{(i)}$ is known to be $\sigma_u^{(i)}$.

We also consider inverse PDE problems given by, 
\begin{equation}
\begin{aligned}
    \mathcal{N}_{\boldsymbol{x}}(u;\boldsymbol{\lambda}) &= f, \quad \boldsymbol{x}\in R^D, \\
    \mathcal{B}_{\boldsymbol{x}}(u;\boldsymbol{\lambda}) & = b, \quad \boldsymbol{x} \in \Gamma,
\end{aligned}
\end{equation}
where $\mathcal{N}_{\boldsymbol{x}}$ is a general differential operator, $D$ represents the number of dimension for a physical domain, $u = u(\bm{x})$ is the solution of the PDE, and $\boldsymbol{\lambda}$ is a vector of unknown parameters in the PDE, $f = f(\bm{x})$ is the forcing term,  $\mathcal{B}_{\boldsymbol{x}}$ is the boundary condition operator, and $\Gamma$ represents the  domain boundary. In addition to $u$, we also have noisy scattered measurements on $f$ and $b$ from sensors. The entire high-fidelity dataset $\mathcal{D}$ can then be expressed as
\begin{equation}
    \begin{aligned}
    \mathcal{D} & = \mathcal{D}_u\cup\mathcal{D}_f\cup\mathcal{D}_b,
        \end{aligned}
\end{equation}
where $ ~\mathcal{D}_f  = \{(\boldsymbol{x}_{f}^{(i)}, ~{f}^{(i)})\}_{i=1}^{N_f}$, and
    $\mathcal{D}_b = \{(\boldsymbol{x}_{b}^{(i)}, ~{b}^{(i)})\}_{i=1}^{N_b}$. 
Similarly, we also assume that the measurements on $f$ and $b$ are independent Gaussian distributions, i.e.,
\begin{equation}
    \begin{aligned}
    {f}^{(i)} &= \bar{f}(\boldsymbol{x}_{f}^{(i)}) + \epsilon_f^{(i)}, \quad i = 1,2...N_f, \\
    {b}^{(i)} &= \bar{b}(\boldsymbol{x}_{b}^{(i)}) + \epsilon_b^{(i)}, \quad i = 1,2...N_b,
    \end{aligned}
\end{equation}
where $\bar{f}$ and $\bar{b}$ are the exact values for $f$ and $b$, respectively, and $\epsilon_f^{(i)}$ and $\epsilon_b^{(i)}$ are independent Gaussian noises with zero mean and known standard deviations, i.e., $\sigma_f^{(i)}$ and $\sigma_b^{(i)}$, respectively.

In the Bayesian framework, we parametrize $u$ with a surrogate model $\tilde{u}(\boldsymbol{x}_u, \tilde{{u}}_L; \boldsymbol{\theta})$, where $\boldsymbol{\theta}$ represents the hyperparameters in the surrogate model (i.e., BNN) and/or unknown parameters in the PDE with a prior distribution $P(\boldsymbol{\theta})$. Consequently, $f$ and $b$ can then be represented by: 
\begin{equation}
\label{eqn:fb}
\begin{aligned}
    \tilde{f}(\boldsymbol{x}_f; \boldsymbol{\theta}) = \mathcal{N}_{\boldsymbol{x}}(\tilde{u}(\boldsymbol{x}_f, \tilde{{u}}_L; \boldsymbol{\theta});\boldsymbol{\lambda}), ~\tilde{b}(\boldsymbol{x}; \boldsymbol{\theta}) = \mathcal{B}_{\boldsymbol{x}_b}(\tilde{u}(\boldsymbol{x}_b, \tilde{{u}}_L; \boldsymbol{\theta}) ;\boldsymbol{\lambda}).
\end{aligned}
\end{equation}
The likelihood for $\mathcal{D}$ can be computed as:
\begin{equation}
\label{eqn:likelihood}
\begin{aligned}
    P(\mathcal{D}|\boldsymbol{\theta}) &= P(\mathcal{D}_u|\boldsymbol{\theta}) P(\mathcal{D}_f|\boldsymbol{\theta}) P(\mathcal{D}_b|\boldsymbol{\theta}), \\
     P(\mathcal{D}_u|\boldsymbol{\theta}) &= \prod_{i=1}^{N_u} \frac{1}{\sqrt{2\pi{\sigma_u^{(i)}}^2}}\exp \left(-\frac{(\tilde{u}(\boldsymbol{x}_{u}^{(i)}; \boldsymbol{\theta}) - \bar{u}^{(i)})^2}{2{\sigma_u^{(i)}}^2}\right), \\
     P(\mathcal{D}_f|\boldsymbol{\theta}) &= \prod_{i=1}^{N_f} \frac{1}{\sqrt{2\pi{\sigma_f^{(i)}}^2}}\exp \left(-\frac{(\tilde{f}(\boldsymbol{x}_{f}^{(i)}; \boldsymbol{\theta}) - \bar{f}^{(i)})^2}{2{\sigma_f^{(i)}}^2}\right), \\
    P(\mathcal{D}_b|\boldsymbol{\theta}) &= \prod_{i=1}^{N_b} \frac{1}{\sqrt{2\pi{\sigma_b^{(i)}}^2}}\exp \left(-\frac{(\tilde{b}(\boldsymbol{x}_{b}^{(i)}; \boldsymbol{\theta}) - \bar{b}^{(i)})^2}{2{\sigma_b^{(i)}}^2}\right). \\
\end{aligned}
\end{equation}
We can then obtain the posterior based on Bayes' rule as:
\begin{equation}
\label{eqn:forwardpost}
\begin{aligned}
P(\boldsymbol{\theta}| \mathcal{D}) = \frac{ P(\mathcal{D}|\boldsymbol{\theta})P(\boldsymbol{\theta})}{P(\mathcal{D})} \simeq P(\mathcal{D}|\boldsymbol{\theta})P(\boldsymbol{\theta}),
\end{aligned}
\end{equation}
in which ``$\simeq$'' represents equality up to a constant. Generally, $P(\mathcal{D})$ is analytically intractable. Here, we employ HMC  to sample from the unnormalized  $P(\boldsymbol{\theta}|\mathcal{D})$ \cite{yang2020b,neal2011mcmc}. Then, we can  obtain predictions on $u$ at any $\bm{x}$ (i.e., $\{\tilde{u}^{(i)}(\boldsymbol{x})\}^M_{i=1}$) based on the posterior samples (i.e., $\{{\boldsymbol{\theta}}^{(i)}\}_{i=1}^M$) from the HMC.  Finally, we can compute the mean as well as the standard deviation for the predictions since the former represents the predictions on  $u(\boldsymbol{x})$ while the latter quantifies the uncertainty. Note that we can also obtain  means and standard deviations for the unknown parameters in the PDEs, i.e., $\bm{\lambda}$, based on the posterior samples in a similar way for $u$.

We note that by $\bm{\lambda}$ we denote a set of unknown constants in the present study. However, in general, $\bm{\lambda}$ can represent an unknown field in this framework and can be represented by another neural network. The interested readers can refer to \cite{yang2020b} for more details.


\subsection{Priors in BNNs}
The surrogate model for $u$, i.e., BNN, is a fully-connected neural network with $L \ge 1$ hidden layers. The input of the neural network is the concatenation of $\bm{x_H}$ and the low-fidelity prediction $\tilde{\bm{u}}_L(\bm{x}_H)$. The output of the $l$-th hidden layer $\bm{z}_{l} \in \mathbb{R}^{N_l}$ ($l = 1,2...N$) is then denoted as
\begin{equation}
    \begin{aligned}
    \bm{z}_{l} &= \phi(\bm{w}_{l-1}\bm{z}_{l-1} + \bm{b}_{l-1}), \quad l = 1, 2...L,\\
    \tilde{u} &= \bm{w}_{L}\bm{z}_{L} + \bm{b}_{L}, 
    \end{aligned}
\end{equation}
where $\bm{w}_l \in \mathbb{R}^{N_{l+1}\times N_l}$ are the weight matrices, $b_l \in \mathbb{R}^{N_{l+1}}$ are the bias vectors, $\phi$ is the nonlinear activation function, which is the hyperbolic tangent function in the present study, and $\bm{z}_0 = (\bm{x}_H, \tilde{\bm{u}}_L(\bm{x}_H)), ~N_0 = N_{x_H} + N_{\tilde{\bm{u}}_L(x_H)}$, and $N_{L+1} = 1$. When using a neural network as a surrogate model, the unknown parameters $\bm{\theta}$ are the concatenation of all the weight matrices and bias vectors.

In  applications of Bayesian neural networks, the independent Gaussian distribution with zero mean for each component of $\bm{\theta}$ is a commonly used prior.  In addition, the entries of $\bm{w}_l$ and $\bm{b}_l$ have the variances $\sigma_{w,l}$ and $\sigma_{b,l}$, respectively, for $l = 0,1...L$ \cite{neal2012bayesian,neal2011mcmc}. In this case, it can be shown that the prior of the function $\tilde{u}(\bm{x})$ is actually a Gaussian process as the width of hidden layers goes to infinity with $\sqrt{N_l}\sigma_{w,l}$ fixed for $l = 1,2...L$ \cite{neal2012bayesian,lee2017deep,pang2019neural}.  In the present study, a BNN with finite width is employed, and we also set the Gaussian distribution as prior for each hyperparameter. The priors for hyperparameters in the same layer are identical. In particular,  $\sigma_{w,l} = \sigma / \sqrt{N}_l$, and $\sigma_{b,l} = 1$, where $l = 0,1...L$, suggesting that the prior distribution is determined by $\sigma$ here.

A good prior is crucial in Bayesian inference, especially for problems with small datasets, e.g., the high-fidelity training data in multi-fidelity modelings. To obtain a proper prior, here we employ the mean-field variational inference (VI) to estimate $\sigma$ by maximizing the marginal likelihood. In particular,  the posterior density of the unknown parameter vector $\bm{\theta} = (\theta_1, \theta_2...\theta_{d_{\bm{\theta}}})$, i.e., $P(\bm{\theta} | \mathcal{D})$, is approximated by another density function $Q(\bm{\theta}; \bm{\zeta})$ parameterized by $\bm{\zeta}$ in the variational inference. A common choice in deep learning  is the mean-field Gaussian approximation \cite{yao2019quality}, i.e., $Q$ is a factorizable Gaussian distribution as follows:
\begin{equation}
\label{eqn:VI_Q}
    Q(\bm{\theta}; \bm{\zeta}) = \prod_{i=1}^{d_{\bm{\theta}}} q(\theta_i; \zeta_{\mu,i}, \zeta_{\rho,i}),
\end{equation}
where $\bm{\zeta} = (\bm{\zeta}_{\mu},\bm{\zeta}_{\rho})$, $\bm{\zeta}_{\mu} = (\zeta_{\mu,1},\zeta_{\mu,2}...\zeta_{\mu,d_{\bm{\theta}}})$, $\bm{\zeta}_{\rho} = (\zeta_{\rho,1},\zeta_{\rho,2}...\zeta_{\rho,d_{\bm{\theta}}})$, and $q(\theta_i; \zeta_{\mu,i}, \zeta_{\rho,i})$ is the density of the one-dimensional Gaussian distribution with mean $\zeta_{\mu,i}$ and standard deviation $\log (1+\exp(\zeta_{\rho,i}))$. 
In this approach, we can tune $\bm{\zeta}$ by maximizing the evidence lower bound (ELBO)  as \cite{blundell2015weight}:
\begin{equation}
    \begin{aligned}
    \arg \max_{\bm{\zeta}, \sigma} \mathcal{L}_{ELBO}  &= \log P(\mathcal{D}) - D_{KL}\left[Q(\bm{\theta};\bm{\zeta})||P(\bm{\theta}|\mathcal{D})\right] \\ \notag
    & \simeq \mathbb{E}_{\bm{\theta}\thicksim Q} [\log Q(\bm{\bm{\theta}}; \bm{\zeta}) - \log P(\bm{\theta}) - \log P(\mathcal{D} | \bm{\theta})],
    \end{aligned}
\end{equation} 
which is equivalent to maximizing the marginal likelihood as well as minimizing the Kullback-Leibler divergence, $D_{KL}$,  between $Q(\bm{\theta};\bm{\zeta})$ and $P(\bm{\theta}|\mathcal{D})$, i.e., $D_{KL}\left[Q(\bm{\theta};\bm{\zeta})||P(\bm{\theta}|\mathcal{D})\right]$. In the present study, we employ the Adam optimizer to obtain the optimal $\bm{\zeta}$ as well as $\sigma$.

The mean-field VI is easy to implement and it is scalable to big data problems by using minibatch training. However, the estimation of posterior distribution (i.e., $Q(\bm{\theta};\bm{\zeta})$) is not quite accurate since the mean-field VI is employed \cite{yang2020b,yao2019quality}. To overcome this drawback, we then employ HMC to estimate the posterior distribution based on the prior provided by the VI. Furthermore, the stochastic HMC can also be used to handle big data in high-dimensional problems. The main steps of the present multi-fidelity Bayesian neural network can be found in Algorithm \ref{alg:mbnns}.

\begin{algorithm}[H]
\caption{Multi-fidelity Bayesian neural network (MBNN)}
\label{alg:mbnns}
\begin{algorithmic}
\Require Low- and high-fidelity noisy measurements and/or PDEs.

\State $\bullet$ Train the first DNN to fit the low-fidelity data using MAP;\;
\State $\bullet$ Train the BNN with high-fidelity training data using mean-fidelity VI to get a proper prior, i.e., $\sigma$;\;
\State $\bullet$ Employ the HMC (or stochastic HMC) to get posterior samples for the hyperparameters in BNN with the priors provided by the mean-fidelity VI;\;
\State $\bullet$ Get predictions for $\{\tilde{u}^{(i)}(\boldsymbol{x})\}^M_{i=1}$) based on the posterior samples from the previous step.
\end{algorithmic}
\end{algorithm}

\section{Results}
\label{sec:results}

We test the performance of the proposed multi-fidelity BNN for diverse applications, including function approximations as well as inverse PDE problems. In addition, we conduct active learning to improve the prediction accuracy based on the predicted uncertainty. For the training of the DNN, we employ the Adam optimizer with an initial learning rate $10^{-3}$, while the number of training steps is set to 50,000 in each case. The weight $\alpha$ for the $\mathcal{L}_2$ regularization is set to  $\sigma^2_{u_L}/N_L$, where $\sigma^2_{u_L}$ and $N_L$ are the variance of the noise and the number for the low-fidelity data, respectively. As for the BNN, we first use the Adam optimizer with an initial learning rate $10^{-3}$ to train it for 200,000 steps, and then we employ 
the HMC with adaptive time step for posterior estimation \cite{yang2020b} with the prior (i.e., $\sigma$) provided by the VI. The number of burn-in steps is set to 10,000,  the initial time step is set as $\delta_t = 0.1$, and the leapfrog step is $L = 50 \delta_t$. Finally, we employ 1,000 posterior samples, i.e., $M = 1,000$, for the computations of the predicted means and standard deviations in each case.

\subsection{Function approximation}
To further explain the method, in this section, we consider approximations of a one-dimensional (1D) and a four-dimensional (4D) function using bi-fidelity data. Both linear and nonlinear correlations between the low- and high-fidelity data are tested.

\subsubsection{1D function approximation}
\label{sec:1dfunc}
\begin{figure}[h]
    \centering
    \subfigure[]{\label{fig:1dfunca}
    \includegraphics[width=0.3\textwidth]{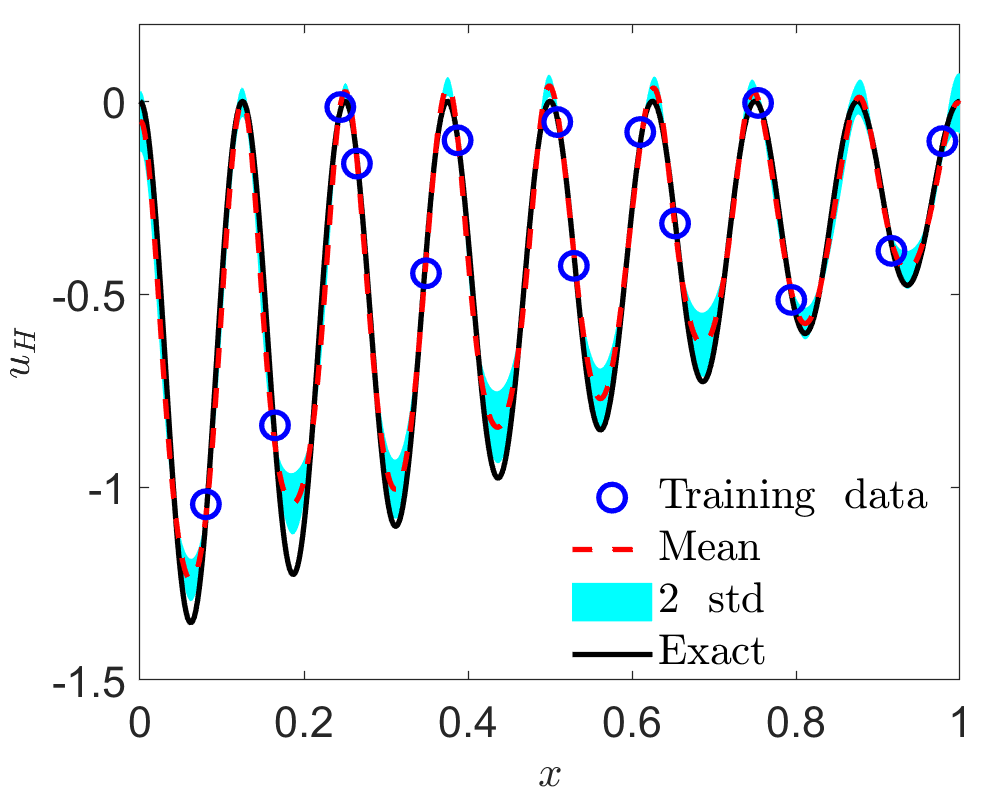}}
    \subfigure[]{\label{fig:1dfuncb}
    \includegraphics[width=0.3\textwidth]{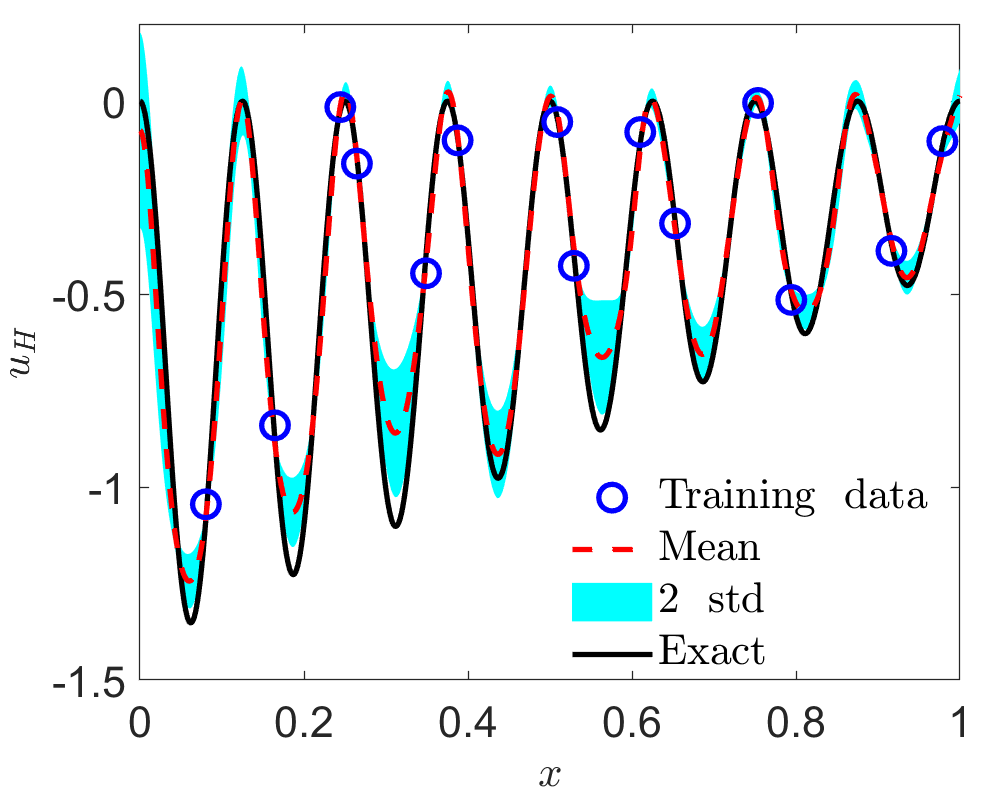}}
    \subfigure[]{\label{fig:1dfuncc}
    \includegraphics[width=0.3\textwidth]{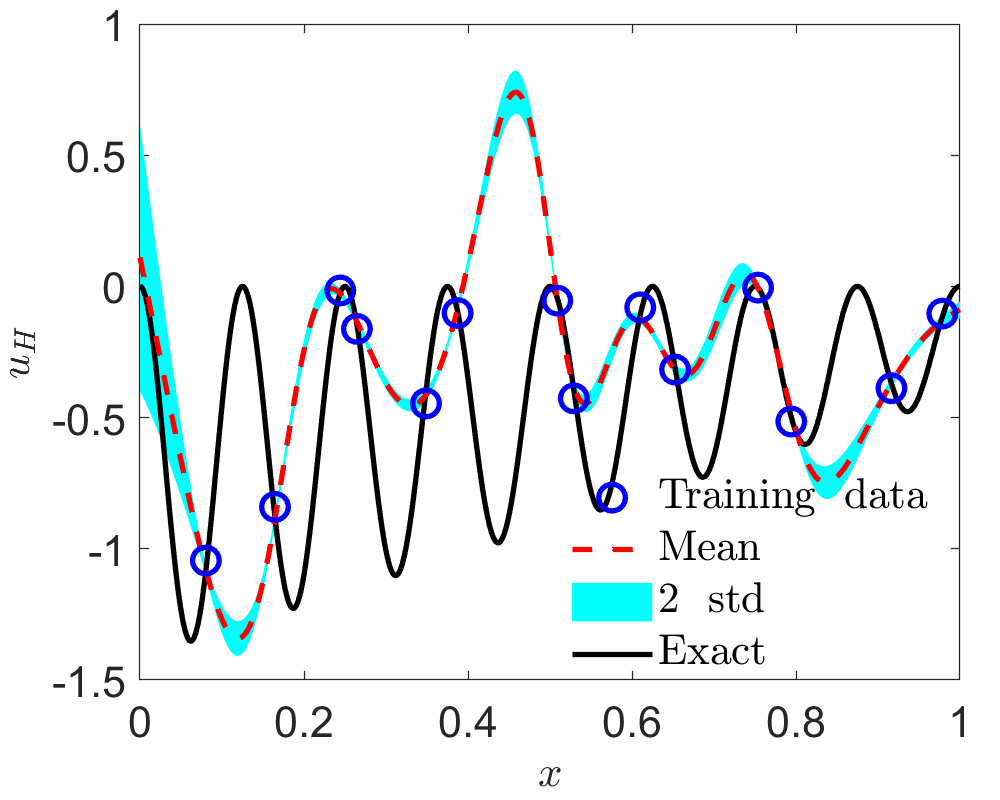}}
    \caption{
    1D function approximation for cases with different noise levels: Predicted means and uncertainties. 
    (a) Multi-fidelity modeling with $\epsilon_L \thicksim 0$, $\sigma = 1.2$. 
    (b) Multi-fidelity modeling with $\epsilon_L \thicksim \mathcal{N}(0, 0.05^2)$, $\sigma = 1.4$. 
    (c) Single-fidelity modeling, $\sigma = 6.5$.
    Blue circles: high-fidelity training data. 2 std: two standard deviations. 
    }
    \label{fig:1dfunc}
\end{figure}

The low- and high-fidelity data are generated using the following two functions \cite{PRDLK17}:
\begin{align}
u_L &= \sin(8x), ~ x \in [0, 1],\\
u_H &= (x - \sqrt{2}) u^2_L,
\end{align}
where $u_L$ and $u_H$ represent the low- and high-fidelity functions, respectively. In addition, we assume that the high-fidelity measurements are noisy. As for the low-fidelity data, we consider two different sources, i.e., (1) ``simulation" data with no noise, and (2) noisy ``measurements". Here, all the noises are assumed to be Gaussian with  zero mean. Specifically, we test two different cases:  (1) $\epsilon_L = 0$, $\epsilon_H \thicksim \mathcal{N}(0, 0.01^2)$; and (2) $\epsilon_L \thicksim \mathcal{N}(0, 0.05^2)$, $\epsilon_H \thicksim \mathcal{N}(0, 0.01^2)$, where $\epsilon_L$ and $\epsilon_H$ denote the noise of the low- and high-fidelity data, respectively. For both cases, we employ 100 uniformly distributed data at the low-fidelity level and 14 high-fidelity measurements, which is the same setup as in \cite{PRDLK17}, as displayed in Fig. \ref{fig:1dfunc}.

In the DNN we employ 2 hidden layers with 20 neurons per layer, while in the BNN we use 1 hidden layer with 50 neurons. The results from the multi-fidelity modeling are illustrated in Fig. \ref{fig:1dfunc}. As for the first case (Fig. \ref{fig:1dfunca}), the multi-fidelity predictions are reasonable: (1) the predicted means are in good agreement with the exact solutions,  and (2) the errors between the predicted means and the exact solutions are mostly bounded by the two standard deviations in the entire domain.  Similar results can also be observed in Fig. \ref{fig:1dfuncb} as we perturb the low-fidelity training data with the Gaussian noise, i.e., $\mathcal{N}(0, 0.05^2)$. However, the predicted uncertainties for the case in Fig. \ref{fig:1dfuncb} are generally larger than those in Fig. \ref{fig:1dfunca}, which can be attributed to the less accurate predictions for the low-fidelity data. 
%
%
 To demonstrate the effectiveness of the multi-fidelity modeling, we further present the results using single-fidelity modeling, i.e., we train the same BNN employed in the multi-fidelity modeling using the high-fidelity data only. As displayed in Fig. \ref{fig:1dfuncc}, the multi-fidelity modeling can significantly enhance the prediction accuracy  compared to the single-fidelity modeling. We note that the results from single-fidelity modeling may be improved by using different architectures for BNN. Here, we employ the same BNN in both single- and multi-fidelity modeling for a fair comparison.

We observe that the priors (i.e., $\sigma$) for the multi-fidelity cases (Figs. \ref{fig:1dfunca} and \ref{fig:1dfuncb}) are slightly different, while they are quite different from the prior in the single-fidelity case (Fig. \ref{fig:1dfuncc}). The difference in the priors of the multi-fidelity cases is attributed to the differences in low-fidelity predictions at $\bm{x}_H$, i.e., $\tilde{\bm{u}}_{L}(\bm{x}_H)$. In addition, the input and output in the single-fidelity modeling are $\bm{x}_H$ and $\bm{u}_H(\bm{x}_H)$, respectively, which is the same as in \cite{yang2020b}.  Considering that different  inputs are employed in the multi-fidelity modeling (Figs. \ref{fig:1dfunca}-\ref{fig:1dfuncb}) and single-fidelity modeling (Fig. \ref{fig:1dfuncc}), the priors, i.e., $\sigma$, in these cases are also quite different. 

\begin{figure}[H]
    \centering
    \subfigure[]{\label{fig:func_03a}
    \includegraphics[width=0.95\textwidth]{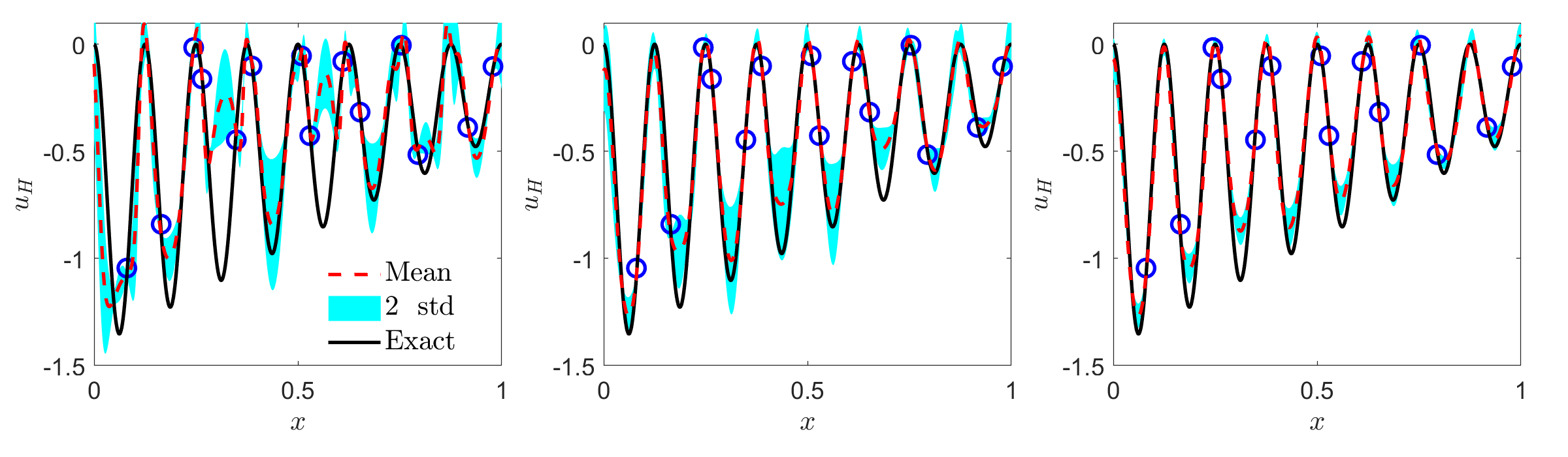}}\\
    \subfigure[]{\label{fig:func_03b}
    \includegraphics[width=0.95\textwidth]{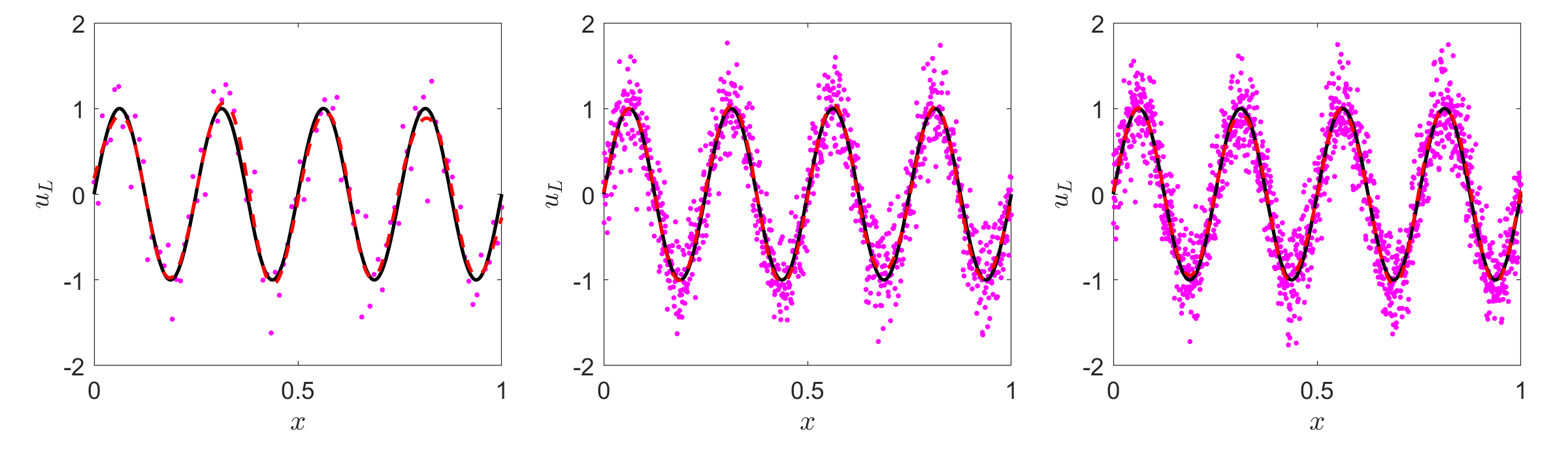}}\\
    \caption{
    Multi-fidelity modeling for 1D function approximations with $\epsilon_L \thicksim \mathcal{N}(0, 0.3^2)$: Different numbers of low-fidelity training data, i.e., 100, 1,000, and 1,500 (from left to right).
    (a) Predicted high-fidelity profiles for different low-fidelity points. From left to right: 100 ($\sigma = 2.2$), 1,000 ($\sigma = 2.2$), and 1,500 ($\sigma = 1.2$). 
    (b) Predicted low-fidelity profiles for different numbers of low-fidelity training data. From left to right: 100, 1,000, and 1,500. 
    }
    \label{fig:func_03}
\end{figure}

We proceed to study the case in which the measurement noise for the low-fidelity data is much larger, i.e., $\epsilon_L \thicksim \mathcal{N}(0, 0.3^2)$. We first test the case with 100 low-fidelity data that are uniformly distributed in $x \in [0, 1]$. The parameters (e.g., optimizer, learning rate, etc.) and architectures for both neural networks are kept the same as the previous case. The low- and high-fidelity predictions are illustrated in Fig. \ref{fig:func_03}; the high-fidelity predictions are not as good as the results in Fig. \ref{fig:1dfunc}, which can be attributed to the inaccurate predictions for the low-fidelity profile (left in Fig. \ref{fig:func_03b}). To enhance the prediction accuracy, we then increase the number of low-fidelity training data, to 1,000, and 1,500. As we can see in the second and third columns of Fig. \ref{fig:func_03b}, the low-fidelity predictions become more accurate. Furthermore, the accuracy for the multi-fidelity modelings (i.e., middle and right panels of Fig. \ref{fig:func_03a}) are also enhanced as compared to the first left in Fig. \ref{fig:func_03a}. In particular, the high-fidelity predictions are quite similar as the results in Fig. \ref{fig:1dfunca} in which the low-fidelity data are noise free. 
It is also interesting to find that the prior (i.e., $\sigma$) for the case with 1,500 low-fidelity training data is the same as in Fig. \ref{fig:1dfunca}, which indicates that the low-fidelity predictions at $\bm{x}_H$ (i.e., $\tilde{\bm{u}}_L(\bm{x}_H)$) are quite similar in these two cases. 
 
All the above results indicate that accurate low-fidelity predictions can enhance the accuracy for multi-fidelity modeling. A possible way to improve the accuracy for low-fidelity predictions is to increase the number of low-fidelity training data as shown in Fig. \ref{fig:func_03}.  

\begin{figure}[H]
    \centering
    \subfigure[]{\label{fig:meana}
    \includegraphics[width=0.45\textwidth]{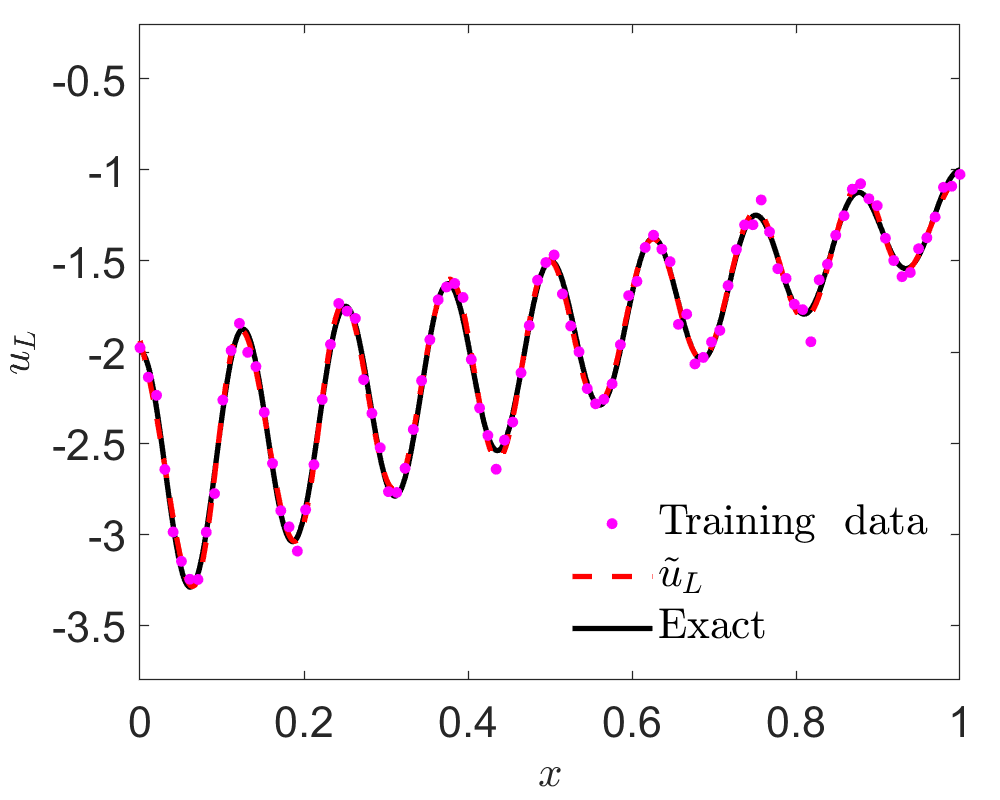}}
     \subfigure[]{\label{fig:meanb}
    \includegraphics[width=0.45\textwidth]{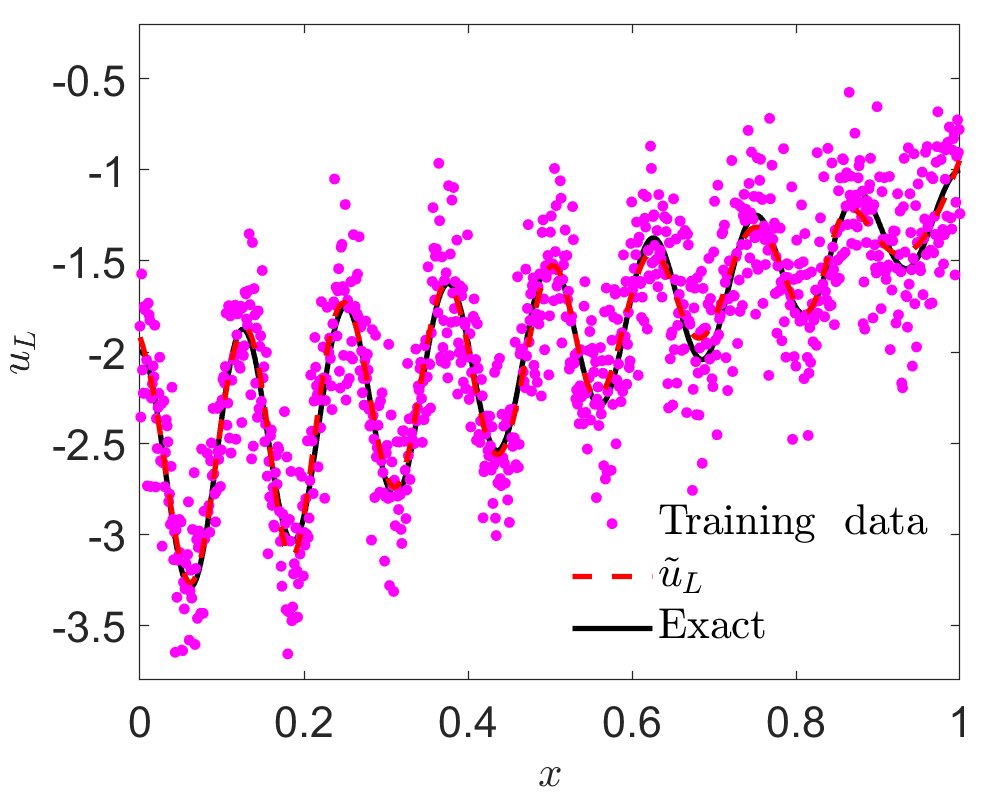}}
    \subfigure[]{\label{fig:meanc}
    \includegraphics[width=0.45\textwidth]{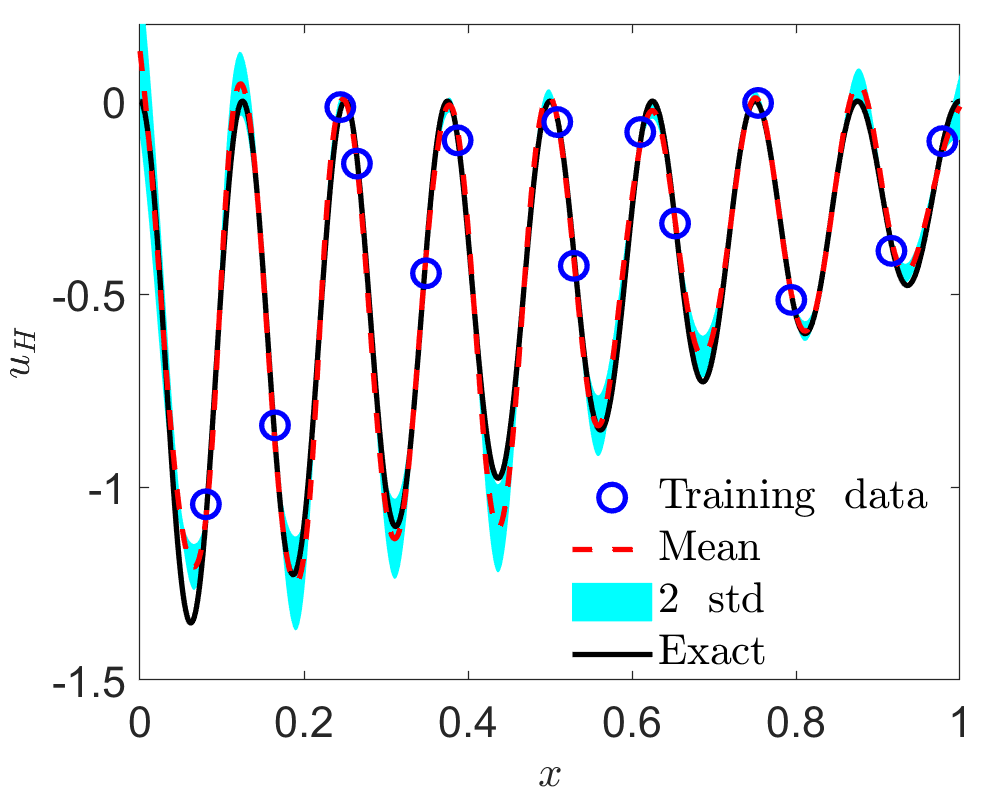}}
     \subfigure[]{\label{fig:meand}
    \includegraphics[width=0.45\textwidth]{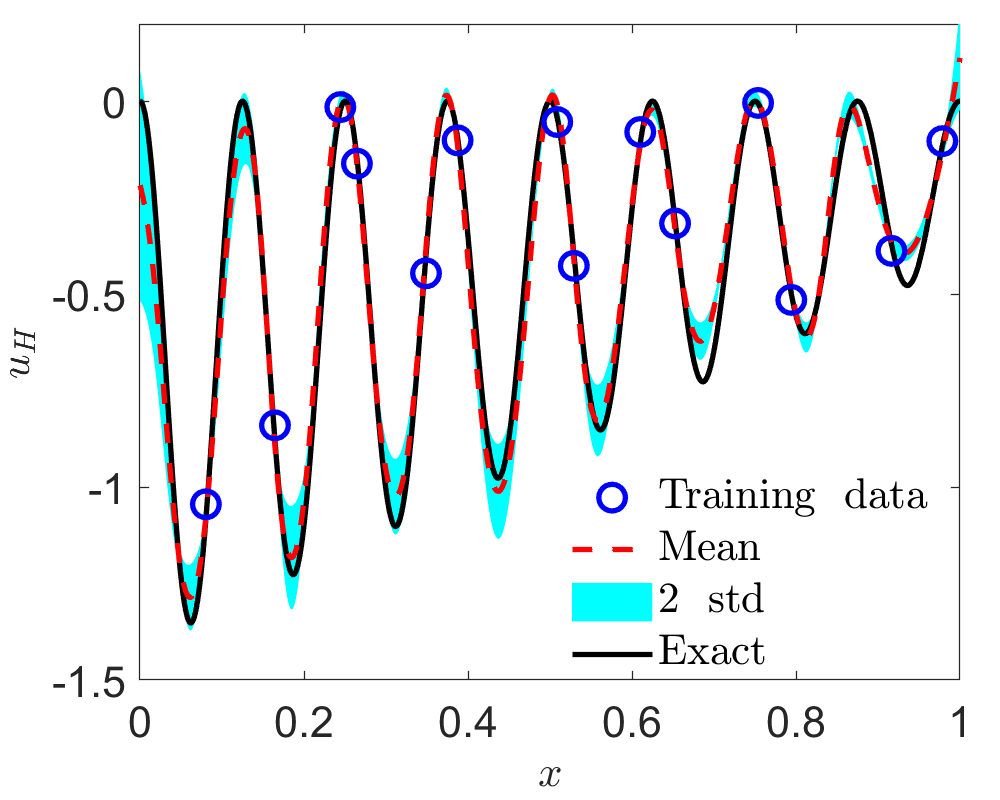}}
    \caption{
    Multi-fidelity modeling for 1D function approximation.  
    (a) Predicted low-fidelity profile for $\epsilon_L \thicksim \mathcal{N}(0, 0.05^2)$. 
    (b) Predicted low-fidelity profile for $\epsilon_L \thicksim \mathcal{N}(0, 0.3^2)$.  
    (c) Predicted means and uncertainties for high-fidelity profiles with $\epsilon_L \thicksim \mathcal{N}(0, 0.05^2)$, $\sigma = 3.1$. 
    (d) Predicted means and uncertainties for high-fidelity profiles with $\epsilon_L \thicksim \mathcal{N}(0, 0.3^2)$, $\sigma = 3.5$. 
    }
    \label{fig:func_mean}
\end{figure}

We further consider a more practical case,  in which both the low- and high-fidelity data are from sensors but with different accuracy. In addition, we assume that there exists a bias between the low- and high-fidelity data, i.e.,  
\begin{align}
    u_L &= (x - \sqrt{2}) \sin^2(8 \pi x) + x - 2, ~ x \in [0, 1],\\
    u_H &= u_L - x + 2.
\end{align}
We test two different noise levels for the low-fidelity data, i.e., $\epsilon_L \thicksim \mathcal{N}(0, 0.05^2)$ and  $\epsilon_L \thicksim \mathcal{N}(0, 0.3^2)$. We employ 100 uniformly distributed low-fidelity training data for the case with $\epsilon_L \thicksim \mathcal{N}(0, 0.05^2)$, and 1,000 uniformly distributed low-fidelity training data for the case with $\epsilon_L \thicksim \mathcal{N}(0, 0.3^2)$; the high-fidelity training data are kept the same as in the previous cases. We see in Fig. \ref{fig:func_mean} that: (1) the predicted means for high-fidelity profiles are in good agreement with the exact solutions for both cases, and (2) the computational errors between the predicted means and exact solutions for $x \in [0, 1]$ are mostly bounded by the two standard deviations.

\subsubsection{4D function approximation}

\begin{figure}[h]
    \centering
    \subfigure[]{\label{fig:4d_funca}
    \includegraphics[width=0.3\textwidth]{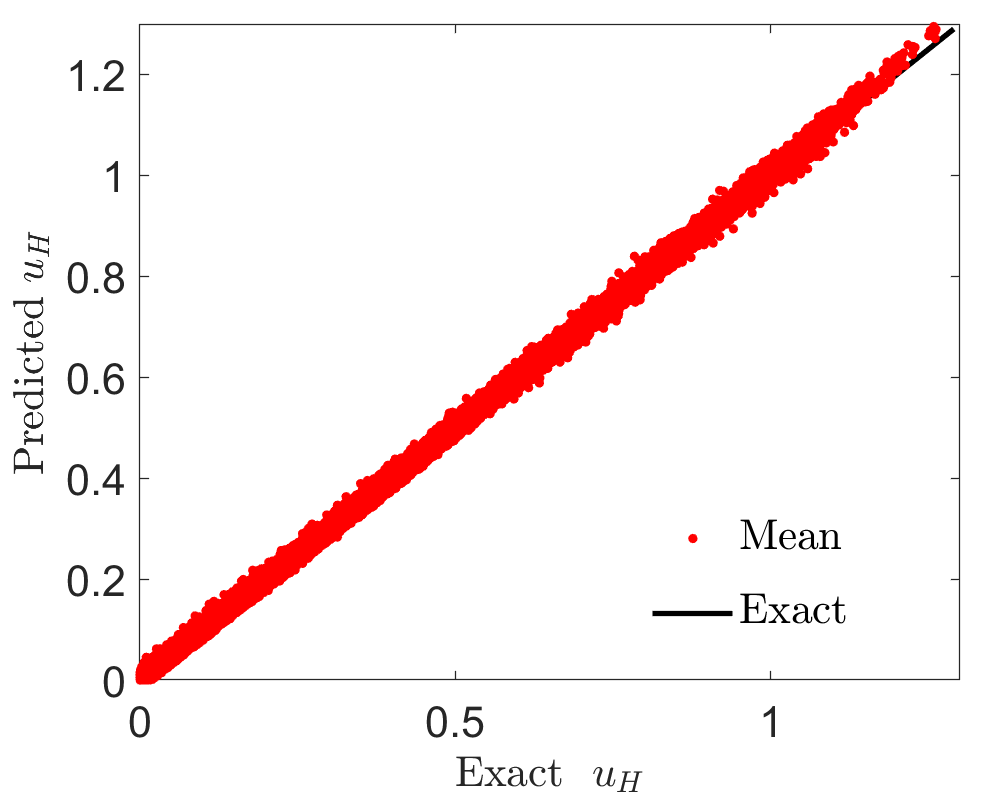}}
    \subfigure[]{\label{fig:4d_funcb}
    \includegraphics[width=0.3\textwidth]{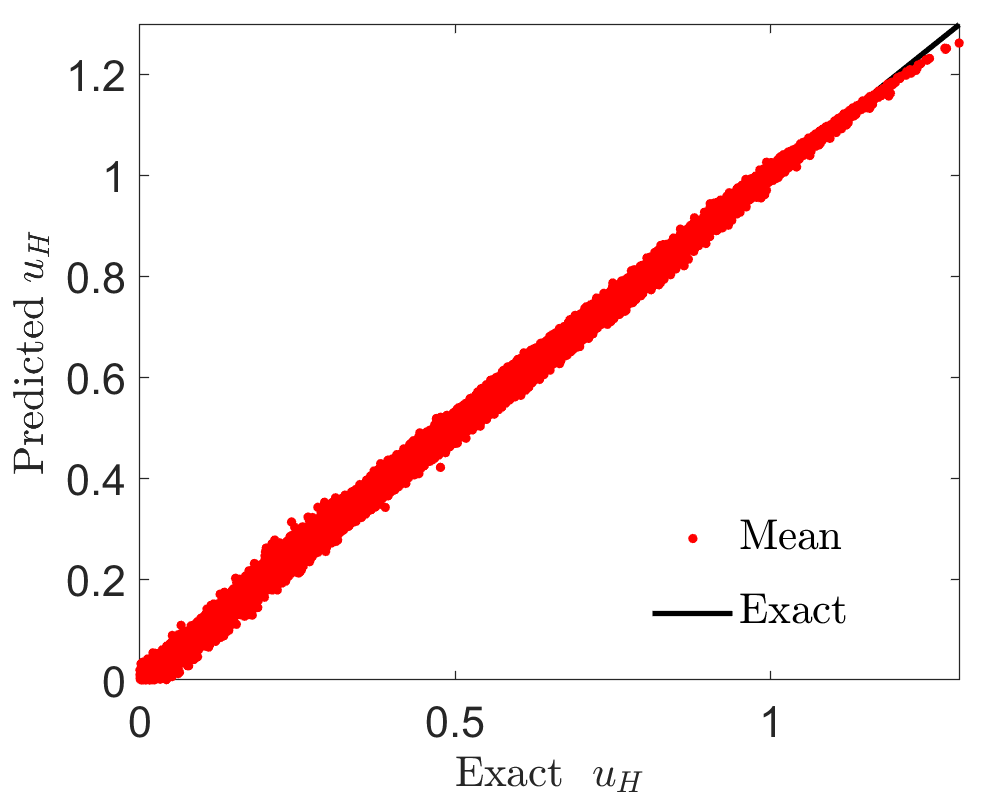}}
    \subfigure[]{\label{fig:4d_funcc}
    \includegraphics[width=0.3\textwidth]{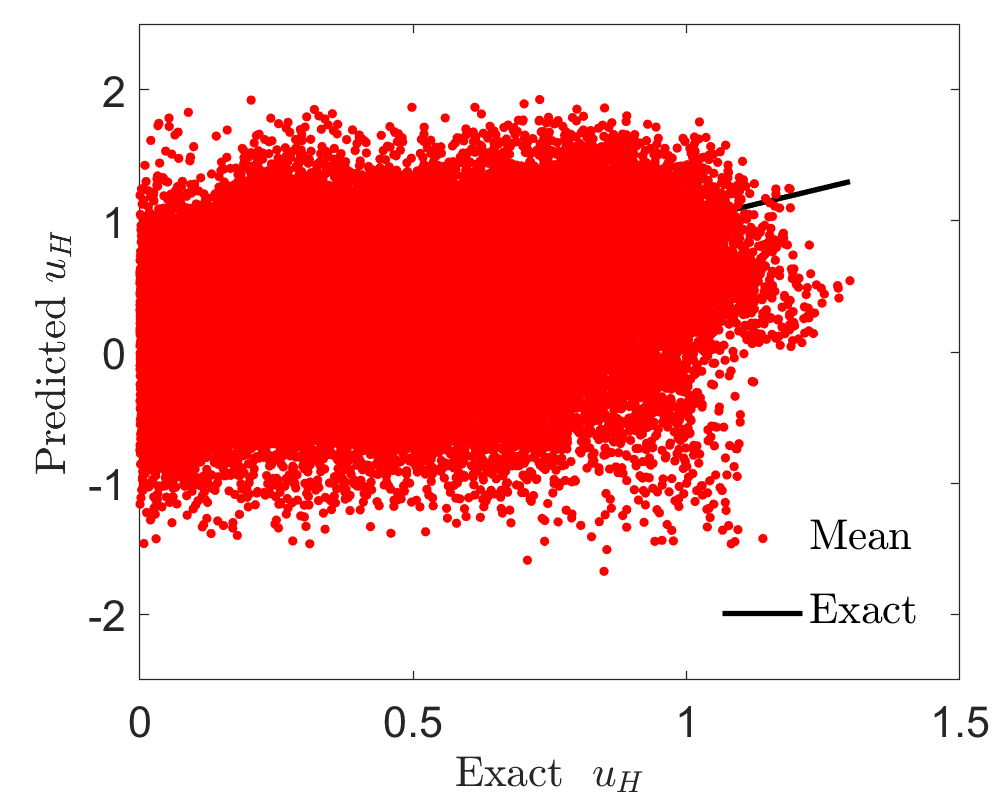}}
    \caption{
    4D function approximation. 
    (a) Predicted high-fidelity data at 100,000 random testing locations with $\epsilon_L \thicksim 0$, $\sigma = 4.2$.
    (b) Predicted high-fidelity data at 100,000 random  testing locations with $\epsilon_L \thicksim \mathcal{N}(0, 0.05^2)$, $\sigma = 4$.
    (c) Single-fidelity predictions at 100,000 random  testing locations, $\sigma = 7.5$.
    }
    \label{fig:4d_func}
\end{figure}

We now consider a four-dimensional function approximation problem using multi-fidelity noisy data. The multi-fidelity training data are generated from the following functions:
\begin{align}
    u_H &= \frac{1}{2}\left[0.1 \exp(x_1 + x_2) - x_4 \sin(12 \pi x_3) + x_3\right], \\
    u_L &= 1.2 u_H - 0.5,
\end{align}
where $x_i \in (0, 1],~ i = 1, 2, 3, 4$. Similarly, we also consider two different sources of low-fidelity data, i.e., (1) ``simulation'' data with no noise, and (2) ``measurements'' with Gaussian noise, i.e., $\mathcal{N}(0, 0.05^2)$. In addition, the measurement errors for the high-fidelity data are still Gaussian but with a smaller variance, i.e., $\mathcal{N}(0, 0.01^2)$.  
In both cases, we employ 25,000 and 150 randomly sampled low- and high-fidelity data for training at the low- and high-fidelity levels, respectively. In addition, the DNN employed for approximating the low-fidelity data has 2 hidden layers with 50 neurons per layer, while for the BNNs, we employ 1 hidden layers with 50 neurons.  Similarly, the same BNN is employed in the single-fidelity modeling for the purpose of comparison.
For validation, we present the high-fidelity predictions at 100,000 random locations. As shown in Fig. \ref{fig:4d_func}, the predicted means from the multi-fidelity BNNs for both cases (Figs. \ref{fig:4d_funca}-\ref{fig:4d_funcb}) are much better than the single-fidelity modeling (Fig. \ref{fig:4d_funcc}).

To evaluate the uncertainty in the test cases, here, we employ the following metric, the prediction interval coverage probability (PICP) \cite{yao2019quality}:
\begin{align}
    \frac{1}{N} \sum^N_{n=1} c_i,
\end{align}
where $N$ is the total number of predictions, and $c_i = 1$ if $u(\bm{x}_n) \in [\tilde{u}_{low}, \tilde{u}_{high}]$, otherwise $C_i = 0$. In addition, $u(\bm{x}_n)$ is the exact value at $\bm{x}_n$, $\tilde{u}_{low} = \tilde{u} - 2 \tilde{u}_{std}$, and $\tilde{u}_{high} = \tilde{u} + 2 \tilde{u}_{std}$, where $\tilde{u}$  and $\tilde{u}_{std}$ denote the predicted mean and  standard deviation for $\tilde{u}$. Ideally, PICP should be close to 1. As displayed in Table \ref{tab:4d_func}, the PICPs for both multi-fidelity cases are greater than  $90 \%$, which means most of the errors in the multi-fidelity modeling are bounded by two standard deviations, 
and the PICPs from the multi-fidelity modeling are much better than in the single-fidelity modeling, which demonstrates the effectiveness of the present multi-fidelity approach. 

\begin{table}[h]
\centering
 \caption{\label{tab:4d_func} 
 4D function approximation with different noise scales: prediction interval coverage probability (PICP). SF: single-fidelity modeling, MF: multi-fidelity modeling. }
 \begin{tabular}{c|ccc}
  \hline \hline
  ~   & MF, $\epsilon_L = 0$   & MF, $\epsilon_L \thicksim \mathcal{N}(0, 0.05^2)  $ & SF\\ \hline
 PICP   & $98.1 \%$ &  $92.0\%$   & $29.8\%$ \\
  \hline \hline
 \end{tabular}
\end{table}

\subsubsection{Multi-fidelity modeling of sea surface temperature (SST)}

\begin{figure}[H]
\centering
\subfigure[]{\label{fig:ssta}
\includegraphics[width = 0.45\textwidth]{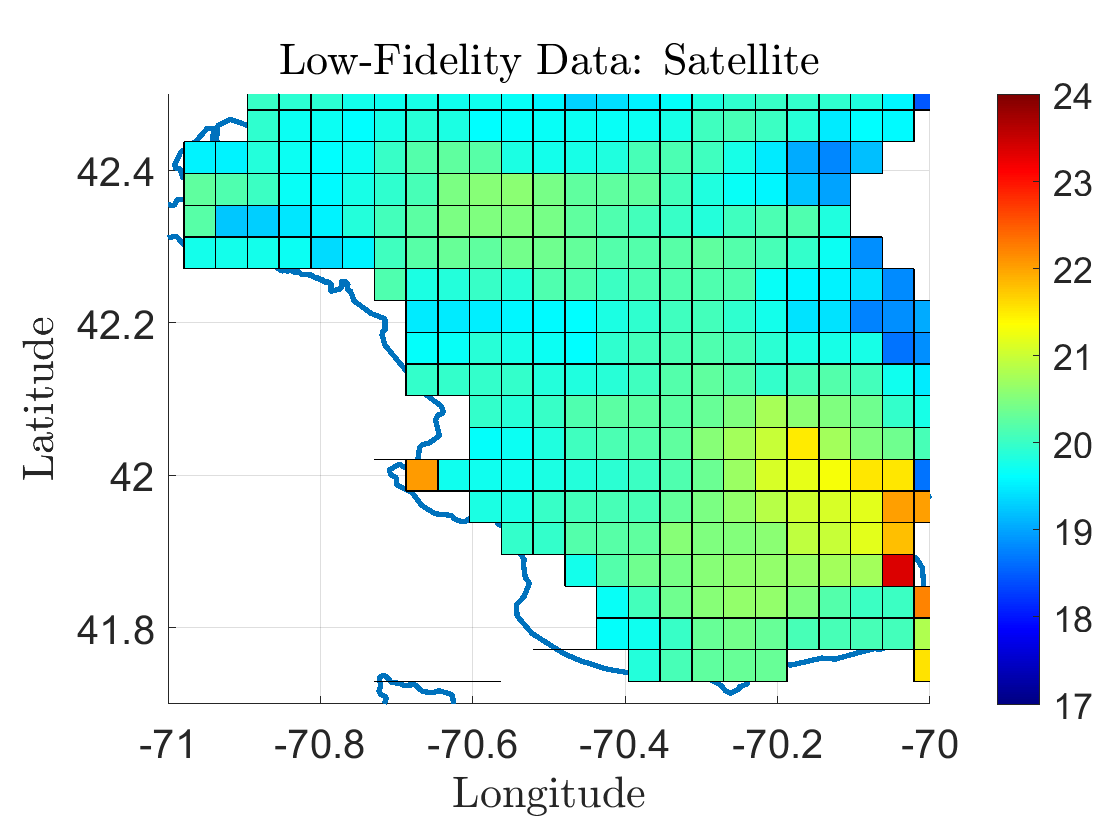}}
\subfigure[]{\label{fig:sstb}
\includegraphics[width = 0.45\textwidth]{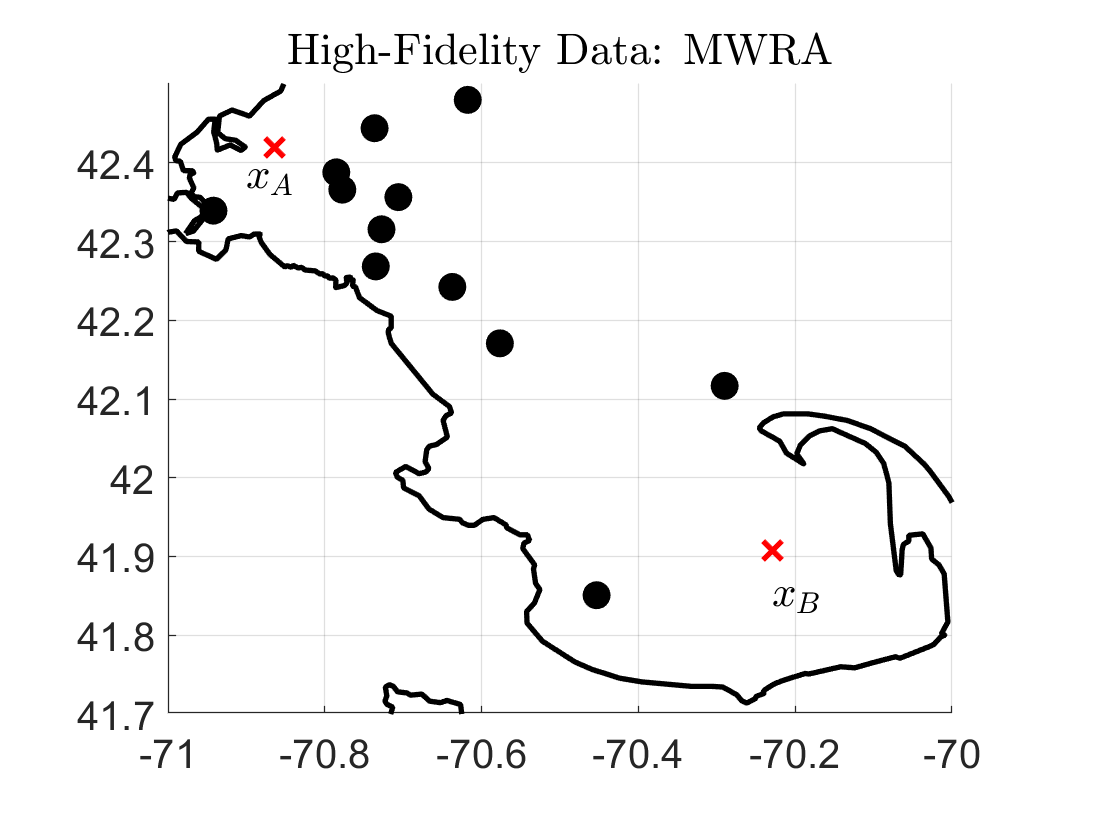}}
\caption{\label{fig:sst_traing}
Training data for sea surface temperature approximation: 
(a) Low-fidelity training data: MODIS. 
(b) High-fidelity data: MWRA. The symbols represent MWRA stations. Black circles: Training data, Red Crosses: Testing data.
}
\end{figure}

We now apply the present model to approximate the sea surface temperature (SST) in the Massachusetts and Cape Cod Bays.  Two sources of data are employed here: (1) Low-fidelity:  The MODerate-resolution Imaging Spectroradiometer (MODIS) Terra on board NASA satellite with the spatial resolution $4 \times 4~ km$ \cite{werdell2013generalized}, and (2) High-fidelity (in situ): The Massachusetts Water Resource Authority (MWRA) \cite{werme2004outfall}. In particular, stations for temperature which are 1$m$ below the sea surface are used as an estimate for seawater surface temperature. More details on the data can be found in \cite{BBDCK20}. The low- and high-fidelity data are for  September 8th, 2015. All the MWRA measurements of different stations are assumed to be taken simultaneously. The locations of MWRA stations are shown in Fig. \ref{fig:sstb}, where  14  measurements are available. Twelve of them are employed as the training data (black circles in Fig. \ref{fig:sstb}), while the others ($x_A: (-70.86, 42.42)$, and $x_B: (-70.23, 41.91)$) (red crosses in Fig. \ref{fig:sstb}) are used to validate the predictions.  

\begin{figure}[H]
\centering
\subfigure[]{\label{fig:ssta}
\includegraphics[width = 0.45\textwidth]{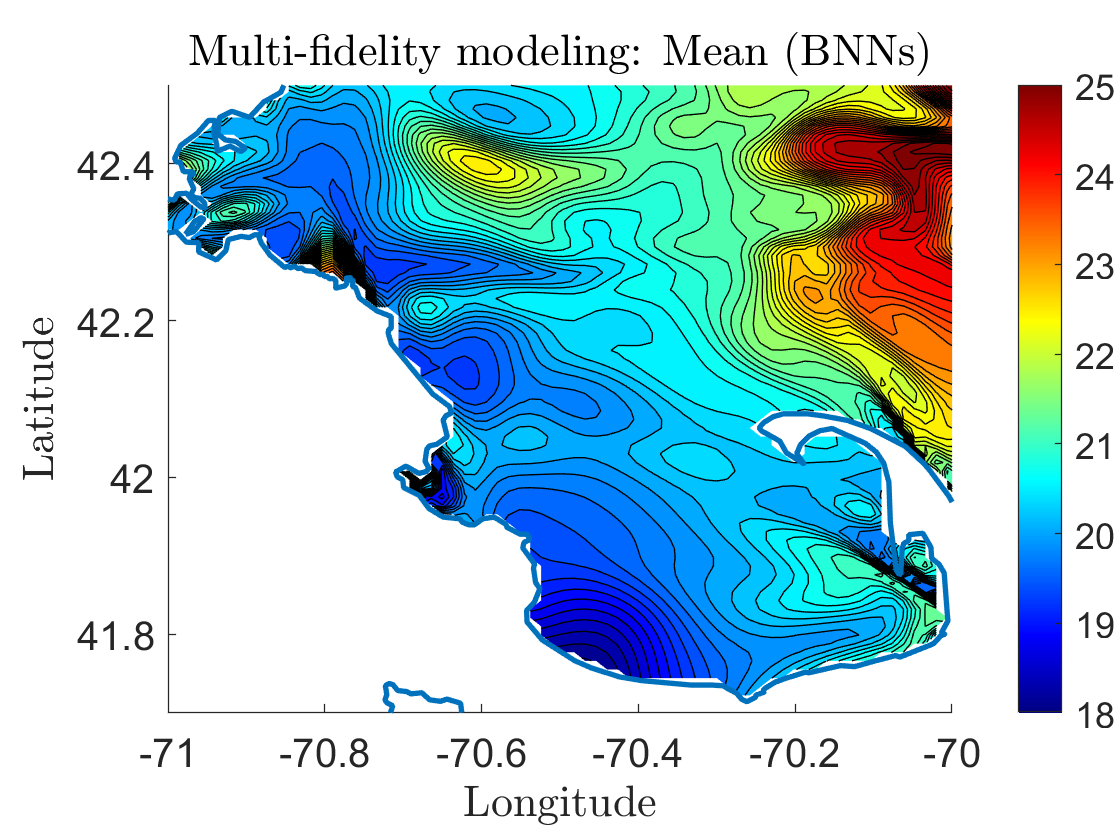}
\includegraphics[width = 0.45\textwidth]{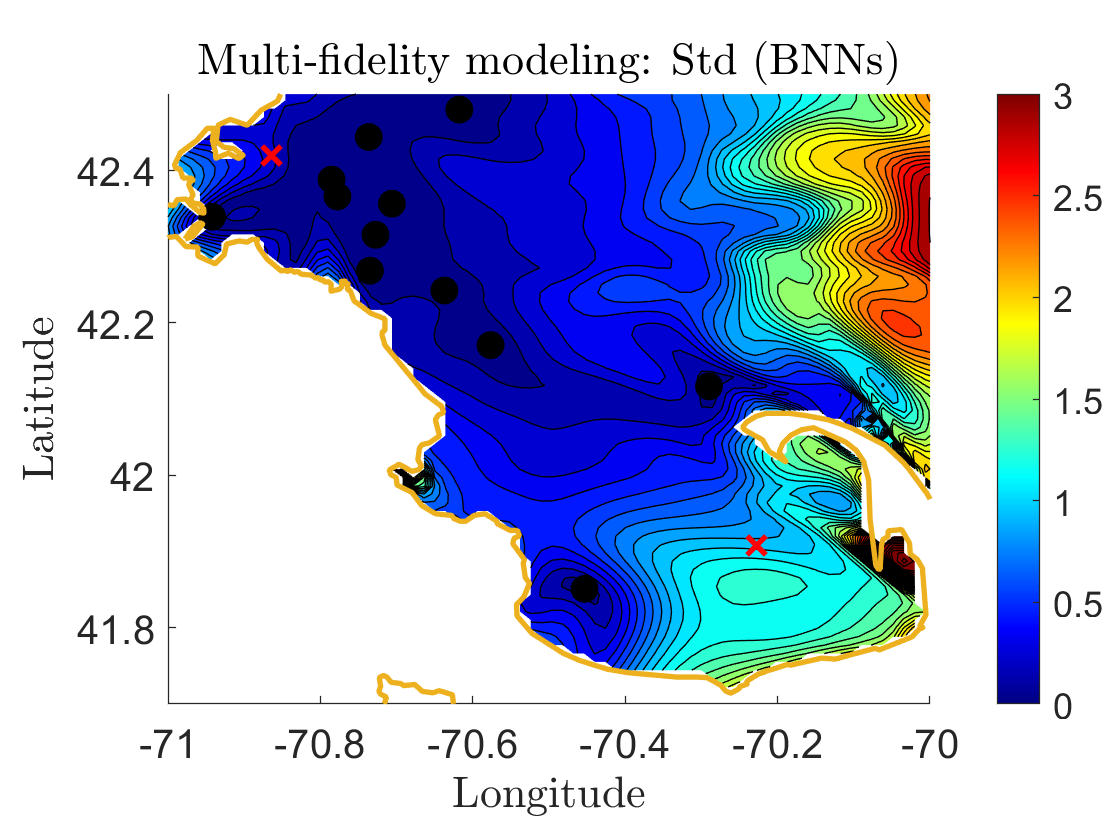}
}
\subfigure[]{\label{fig:sstb}
\includegraphics[width = 0.45\textwidth]{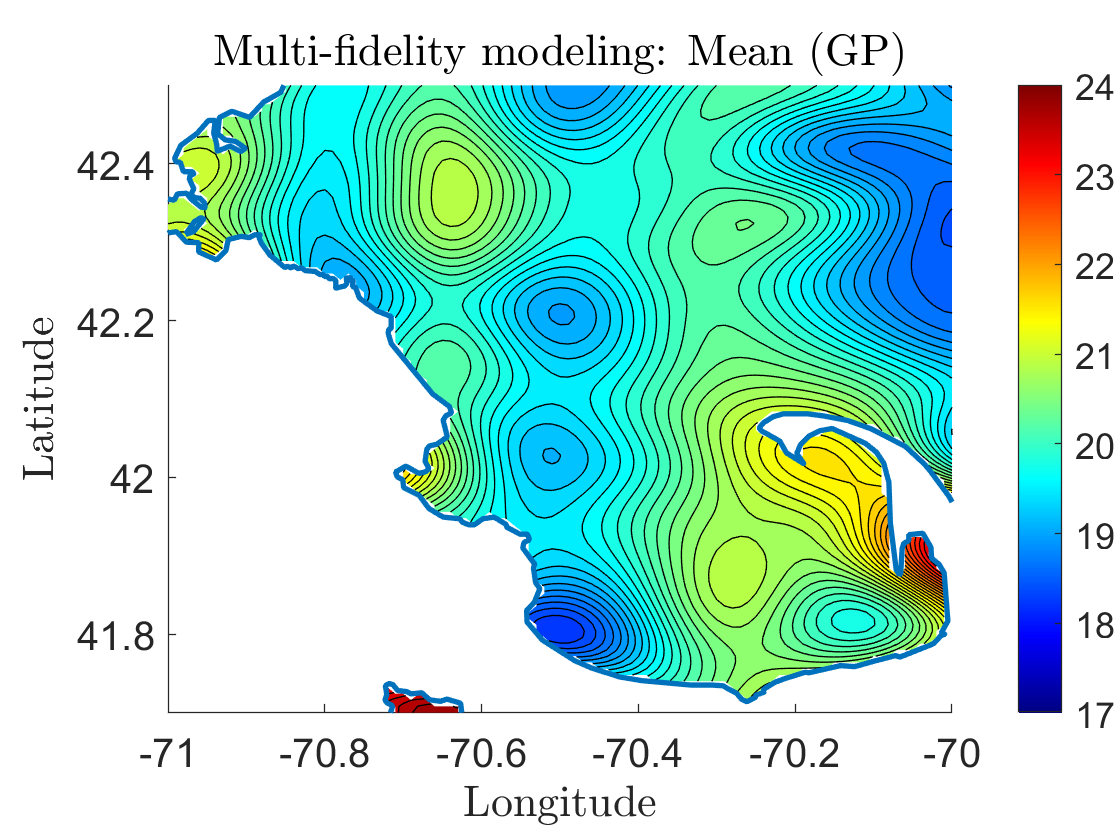}
\includegraphics[width = 0.45\textwidth]{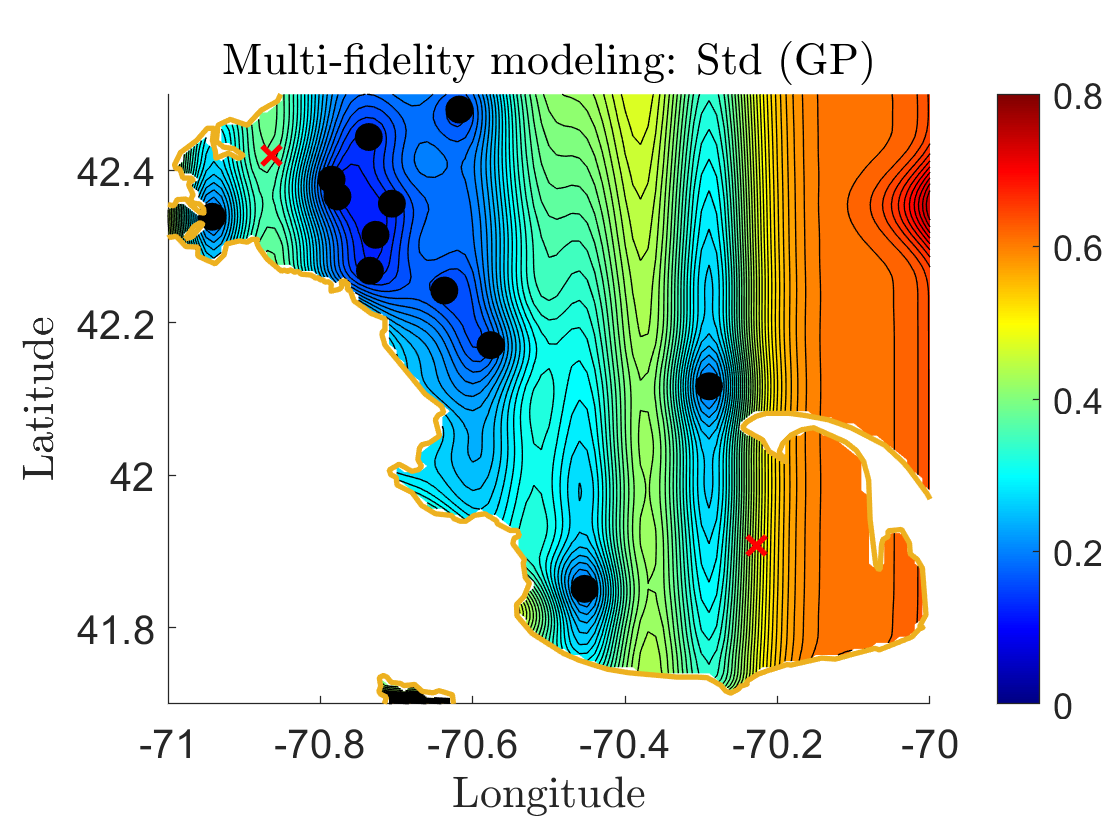}
}
\subfigure[]{\label{fig:sstc}
\includegraphics[width = 0.45\textwidth]{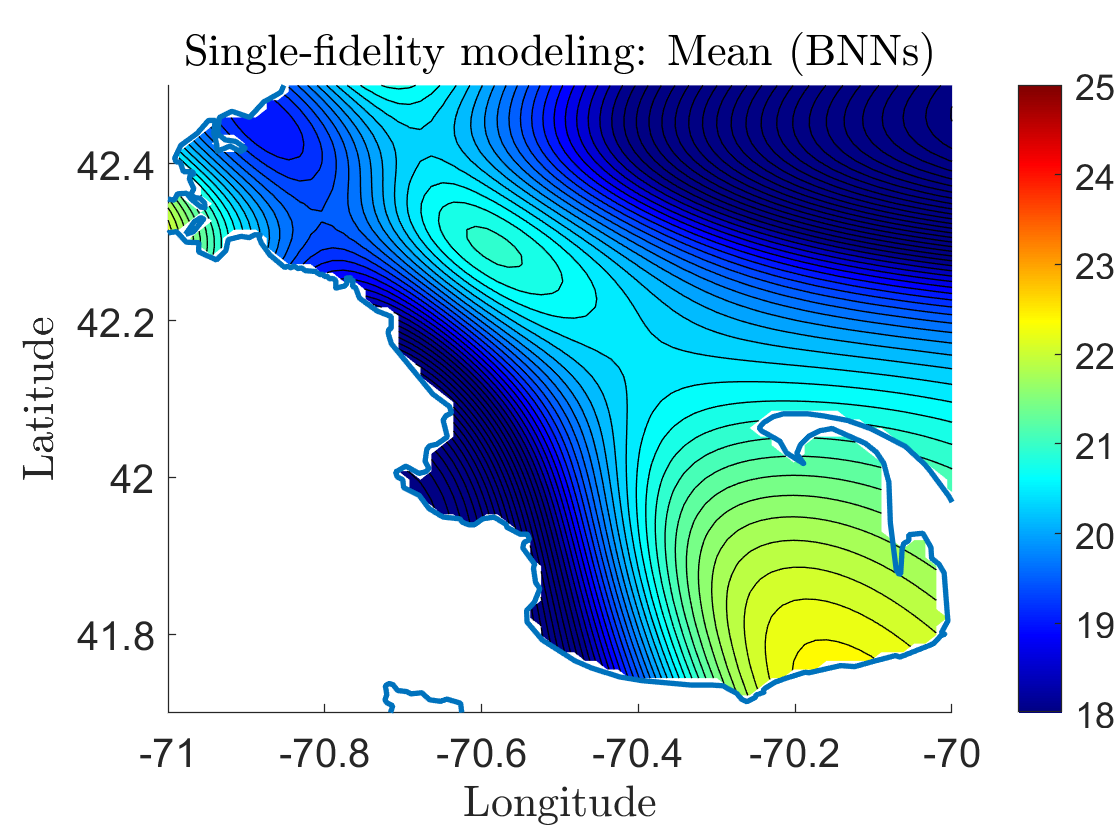}
\includegraphics[width = 0.45\textwidth]{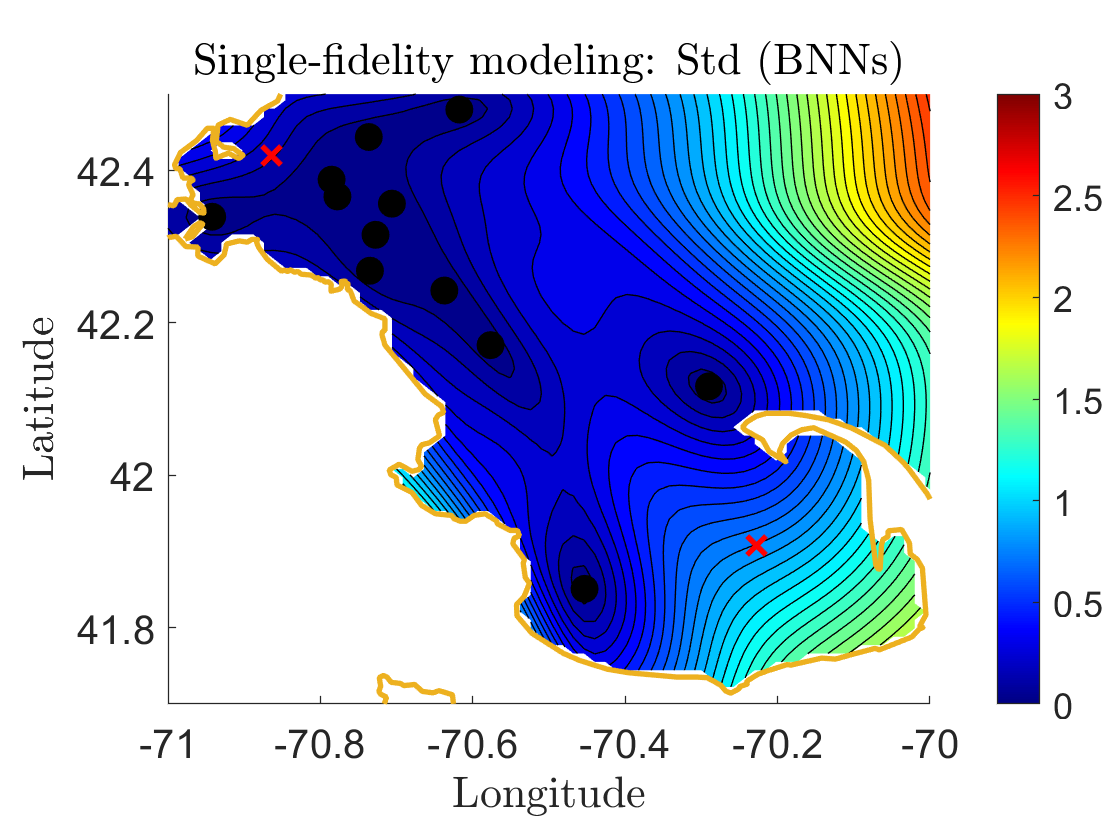}
}

\caption{\label{fig:sst}
Approximation of sea surface temperature from multi-fidelity data: 
(a) Predicted means and uncertainties from multi-fidelity BNNs with $\sigma = 1.8$.
(b) Predicted means and uncertainties from multi-fidelity GPR. 
(c) Predicted means and uncertainties from single-fidelity  BNNs with $\sigma = 1.8$.
}
\end{figure}

In multi-fidelity modeling, we employ 2 hidden layers with 40 neurons per layer for the low-fidelity NN, and only 1 hidden layer with 50 neurons in the BNN.  The predicted  temperature field and the uncertainty map using BNN are shown in Fig. \ref{fig:ssta}. It is clear that the uncertainty is near zero close to MWRA stations (high-fidelity measurements) while the uncertainty is much higher in the north east of the map, where no MWRA in situ measurements are available. We also show similar results obtained from GPR multi-fidelity model and single-fidelity BNNs in Figs. \ref{fig:sstb}  and \ref{fig:sstc}, respectively. Both BNNs and GPR multi-fidelity models show finer structure SST variations as they both utilize the high resolution (in comparison with scarce MWRA measurements)  satellite data, while the single-fidelity BNN captures the large structures in the SST as it only utilizes scarce (12) MWRA measurements.   Furthermore, we display the predicted SSTs at the two testing locations, i.e., $x_A$ and $x_B$, in Table \ref{tab:sst}, as well as the results from single-fidelity modeling. We observe that the predicted means at the two testing locations from these two methods have comparable accuracy. However, the multi-fidelity modeling provides more accurate prediction for $x_A$ than the single-fidelity modeling. While this demonstration shows that BNN performs comparably to GPR multi-fidelity, it is important to note that training the GPR multi-fidelity for large datasets, for example annual space-time SST models, can be computationally prohibitive, whereas training the BNN can be done in a scalable manner for large training data.

\begin{table}[h]
\centering
 \caption{\label{tab:sst} Single- and multi-fidelity predictions for the temperature at the testing locations (mean $\pm$ one standard deviation). MF-BNNs: multi-fidelity BNNs, MF-GP: multi-fidelity GP,  SF-BNNs: single-fidelity BNNs.}
 \begin{tabular}{c|cccc}
  \hline \hline
  ~ & MWRA  & MF-BNNs & MF-GP   & SF-BNNs   \\ \hline
  $x_A $ & 19.72   & 20.12 $\pm$ 0.27  & 20.10 $\pm$ 0.38 &  19.13 $\pm$ 0.18   \\
 $x_B $  & 21.27  & 20.52 $\pm$ 1.01 & 20.84 $\pm$ 0.46 &  21.94 $\pm$ 0.81    \\
  \hline \hline
 \end{tabular}
\end{table}

\subsection{Data-driven solutions for inverse PDE problems}
 Next, we employ PINNs attached to BNN (see Fig. \ref{fig:bnn}) to test our method using PDE-based problems.

\subsubsection{1D inverse PDE problem}
\label{sec:1Dpde}
We first consider the following 1D nonlinear problem, which is described by \begin{align}\label{eq:1dinv}
    \frac{1}{192 \pi^2}u_{xx} -  \frac{k}{24 \pi} u u_x &= f(x), ~ x \in [0, 1],
\end{align}
where $k = 1$ is a constant. The exact solution to this problem is expressed as $u(x) = (x - \sqrt{2})\sin^2(8 \pi x)$, and $f(x)$ can  be derived from Eq. \eqref{eq:1dinv}. Here, we assume that $k$ is unknown, and the objective is to identify the value of $k$ given a limited number of measurements for $u(x)$ and $f(x)$.

As for measurements, we assume that we have high-fidelity sensors for $u(x)$ and  $f(x)$, respectively. In particular, we have the same number of sensors for both $u$ and $f$, i.e., 10, which are randomly distributed in the computational domain,  and we have two additional sensors for $u(x)$, which are at the two boundaries to provide Dirichlet boundary conditions. Finally, we assume that the measurement errors for $u$ and $f$ follow Gaussian distributions, i.e., $\epsilon_u \thicksim \mathcal{N}(0, 0.01^2)$, and $\epsilon_f \thicksim \mathcal{N}(0, 0.01^2)$.  

\begin{figure}[H]
    \centering
    \subfigure[]{\label{fig:1d_inversea}
    \includegraphics[width=0.9\textwidth]{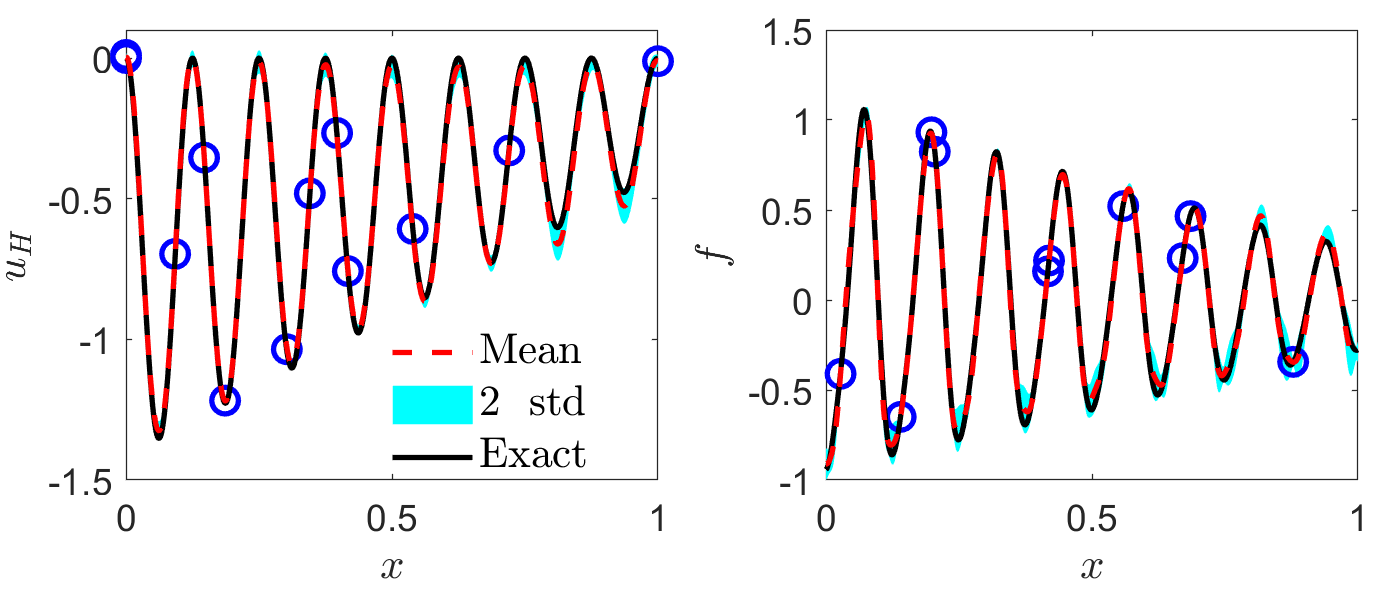}}
    \subfigure[]{\label{fig:1d_inverseb}
    \includegraphics[width=0.9\textwidth]{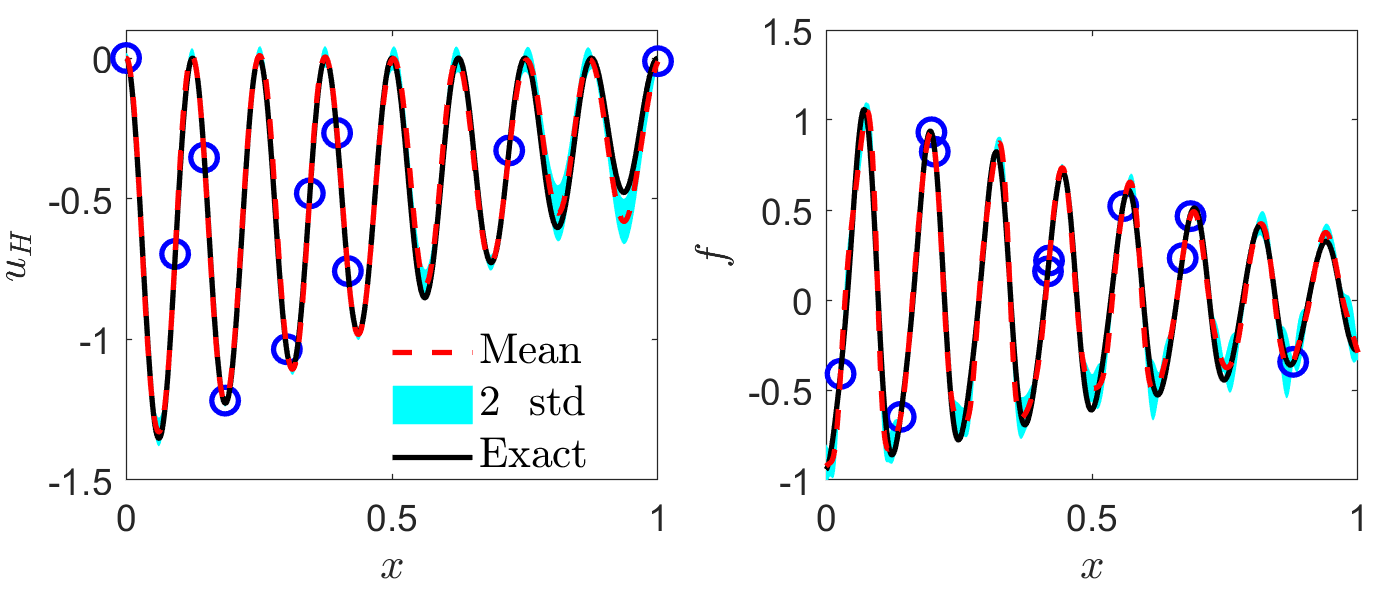}}
    \subfigure[]{\label{fig:1d_inversec}
    \includegraphics[width=0.9\textwidth]{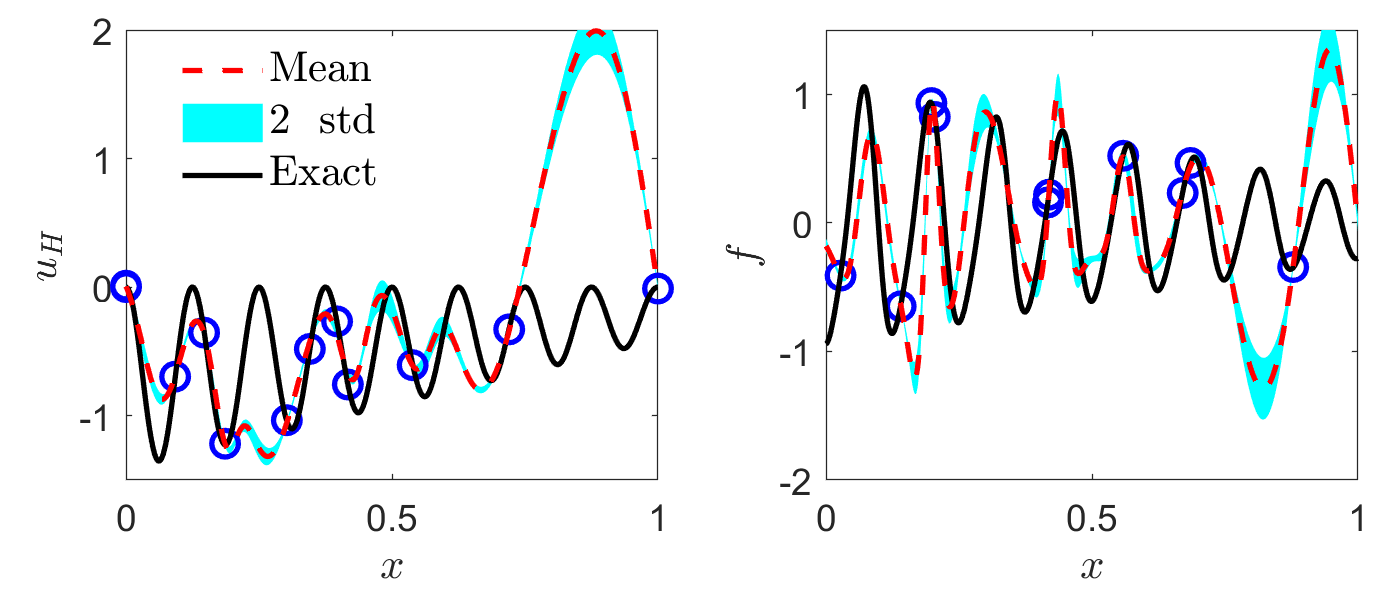}}
    \caption{
    1D inverse PDE problem. 
    (a) Predicted means and uncertainties for $u$ and $f$ using multi-fidelity data: $\epsilon_{u_L} = 0$, and $\sigma = 1.4$. 
    (b) Predicted means and uncertainties for $u$ and $f$ using multi-fidelity data: $\epsilon_{u_L} \thicksim \mathcal{N}(0, 0.05^2)$), and $\sigma = 1.4$.
     (c) Predicted means and uncertainties for $u$ and $f$ using  high-fidelity data only with $\sigma = 4.6$.
    Blue circles: high-fidelity training data.
    }
    \label{fig:1d_inverse}
\end{figure}

\begin{table}[h]
\centering
{\footnotesize
\begin{tabular}{c|cccc}
\hline \hline
     & {MF, $\epsilon_{u_L} = 0$} & {MF, $\epsilon_{u_L} \thicksim \mathcal{N}(0, 0.05^2)$} & {SF}  \\
  \hline
  Mean  & 1.006 & 1.031  & 4.128 \\
   Std    & {\footnotesize $4.20 \times 10^{-2}$} & {\footnotesize $5.08 \times 10^{-2}$} &   {\footnotesize $1.63 \times 10^{-1}$}\\ 
  \hline \hline
\end{tabular}
}
\caption{
1D inverse PDE problem: Predicted mean and standard deviation for $k$, where $k = 1$ is the exact solution. MF, $\epsilon_{u_L} = 0$: multi-fidelity modeling with $\epsilon_{u_L} = 0$; MF, $\epsilon_{u_L} \thicksim \mathcal{N}(0, 0.05^2)$: multi-fidelity modeling with $\epsilon_{u_L} \thicksim \mathcal{N}(0, 0.05^2)$; SF: single-fidelity modeling.
}
\label{tab:1d_kpred}
\end{table}

To obtain the low-fidelity training data, we can either solve Eq. \eqref{eq:1dinv} numerically or analytically with a guessed $k$ and $f$ based on our  prior knowledge, which is similar to the work in \cite{meng2020composite}, or directly measure $u$ using low-fidelity sensors. Here we simply assume that the exact  low-fidelity function is $u_L(x) = \sin(8 \pi x)$, which is the same as in Sec. \ref{sec:1dfunc}. Similarly, we consider the following two different low-fidelity data: (1) simulation data with $\epsilon_{u_L} = 0$, and (2) measurements with Gaussian noise, $\epsilon_{u_L} \thicksim \mathcal{N}(0, 0.05^2)$. In both cases, 500 uniformly distributed low-fidelity samples are employed. In the DNN for the low-fidelity data, we employ 2 hidden layers with 20 neurons per layers, while in the BNN we use 1 hidden layer with 50 neurons. To obtain a proper prior distribution for $k$, we can either follow the same way for obtaining the priors of the hyperparameters in BNNs, i.e.,  maximizing the ELBO using VI, or set it directly based on the prior physical knowledge. Here we employ the latter approach for the prior of $k$ for simplicity, i.e.,  a standard normal distribution here. The results from the multi-fidelity modeling are illustrated in Figs. \ref{fig:1d_inverseb}-\ref{fig:1d_inversec}. As shown, the predicted means for $u$ and $f$ agree with the exact solutions well for both cases, which can also be reflected by the small predicted uncertainties. In addition, the predicted uncertainties for both $u$ and $f$ for the case in Fig. \ref{fig:1d_inverseb} are larger than those in Fig. \ref{fig:1d_inversea}, which is reasonable due to the noise scale for the low-fidelity data in the former case, which is larger than in the latter case. Note that similar results have also been observed in Sec. \ref{sec:1dfunc}, and will not be discussed in detail here. Furthermore, we  present the results from single-fidelity modeling, i.e., using the same BNNs and train them with the high-fidelity training data only, which is the same as in the Bayesian physics-informed neural networks (B-PINN) \cite{yang2020b}. As shown in Fig. \ref{fig:1d_inversec}, the multi-fidelity results are more accurate than the single-fidelity solutions. 

Finally, the predicted $k$ for the three test cases in Fig. \ref{fig:1d_inverse} are displayed in Table \ref{tab:1d_kpred}, which demonstrates: (1) the predicted means in both multi-fidelity cases are in good agreement with the exact solutions, (2) the computational error increases with the increasing noise scale in low-fidelity data, which is consistent with the results in Figs. \ref{fig:1d_inversea}-\ref{fig:1d_inverseb}, (3) the computational errors in both multi-fidelity cases have the similar order as the noise scale, and they are all bounded by the one standard deviation, and (4) the multi-fidelity modeling outperforms the single-fidelity modeling with the aid of low-fidelity data.

\subsubsection{2D inverse PDE problem}
\label{sec:2Dpde}
We proceed to consider a 2D steady-state diffusion-reaction system as
\begin{align}\label{eq:2dpde}
    \lambda \left(\partial^2_x u + \partial^2_y u\right) - k u^2 = f, ~ x,~y~ \in [-1, 1], 
\end{align}
where $u$ represents the concentration, $\lambda = 0.01$ is the diffusion coefficient, $k$ is the reaction rate, and $f$ is the source term. The exact solution to Eq. \eqref{eq:2dpde} is $u = \sin(2 \pi x) \sin(2 \pi y)$, while $f$ can be derived from the exact solution for Eq. \eqref{eq:2dpde}. Here, we assume that we have a limited number of high-fidelity sensors of $u$ and $f$ rather than exact expressions for them. Similarly, we also assume that $k$ is an unknown constant. The objective is then to predict $u$ and $f$ in the entire computational domain as well as to identify $k$.

Similarly, we assume that we have lots of low-fidelity observations for $u$, which are generated from $u_L = 0.8 u + 0.2$ here. It is worth mentioning that the low-fidelity data can be obtained from numerical simulations with a guessed $f$ or measurements with lower accuracy, which are similar as the assumption in Sec. \ref{sec:1Dpde}. Therefore, we again consider two scenarios for the low-fidelity data, i.e., (1) simulation data with no noise, and (2) low-fidelity measurements with Gaussian noise $\epsilon_{u_L} \thicksim \mathcal{N}(0, 0.05^2)$. In both cases, we assume that the noise for the high-fidelity measurements is the same, i.e.,  $\epsilon_{u} \thicksim \mathcal{N}(0, 0.01^2)$, and $\epsilon_{f} \thicksim \mathcal{N}(0, 0.01^2)$.   

\begin{figure}[H]
    \centering
    \subfigure[]{\label{fig:2d_inversea}
    \includegraphics[width=0.9\textwidth]{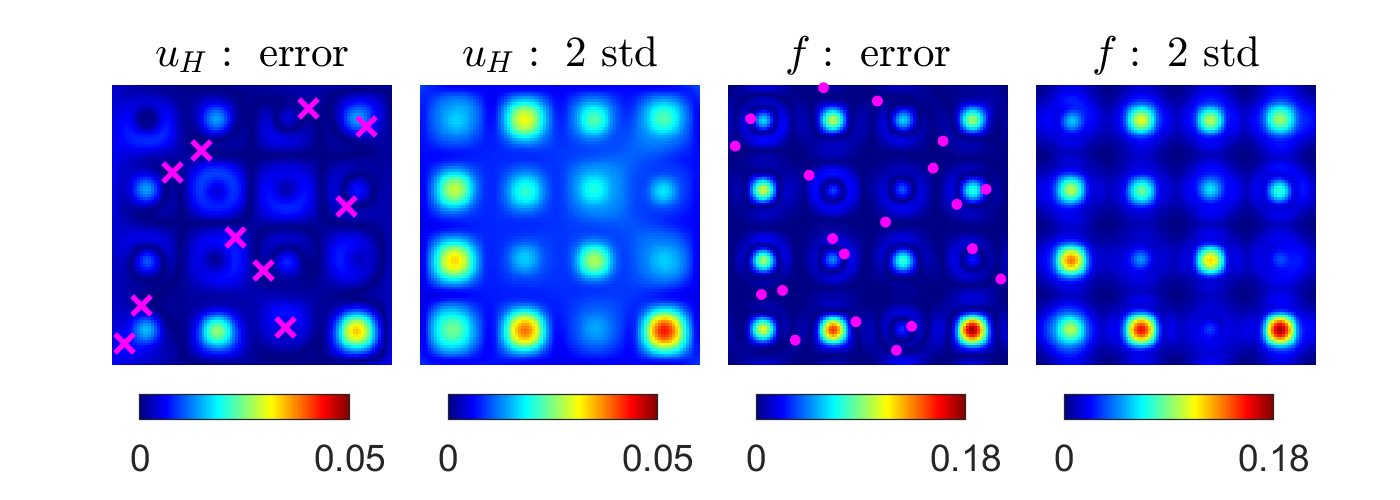}}
    \subfigure[]{\label{fig:2d_inverseb}
    \includegraphics[width=0.9\textwidth]{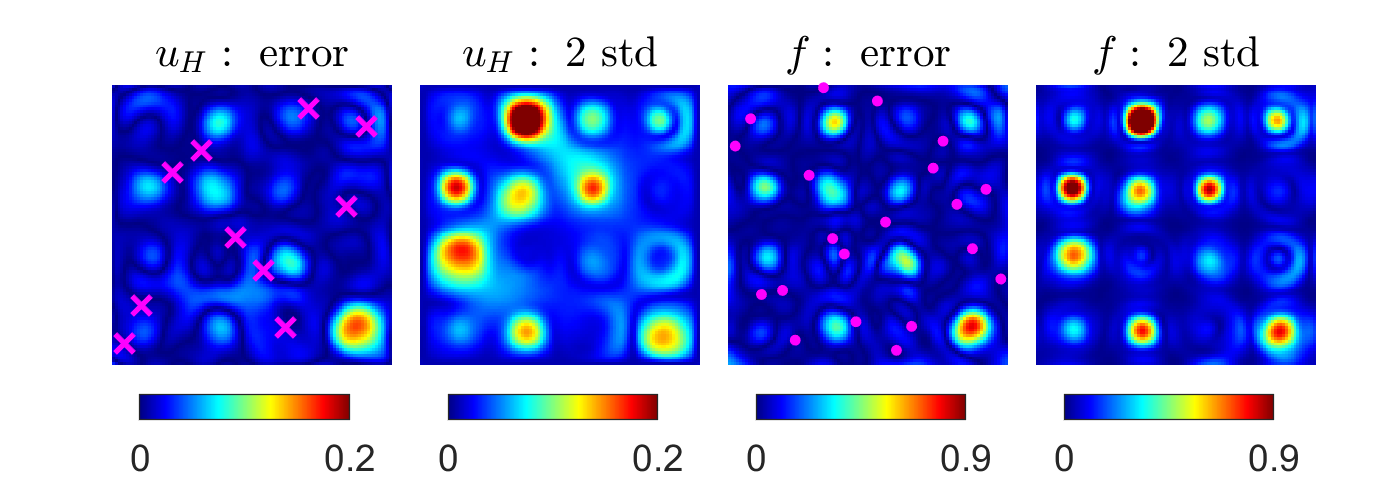}}
    \subfigure[]{\label{fig:2d_inversec}
    \includegraphics[width=0.9\textwidth]{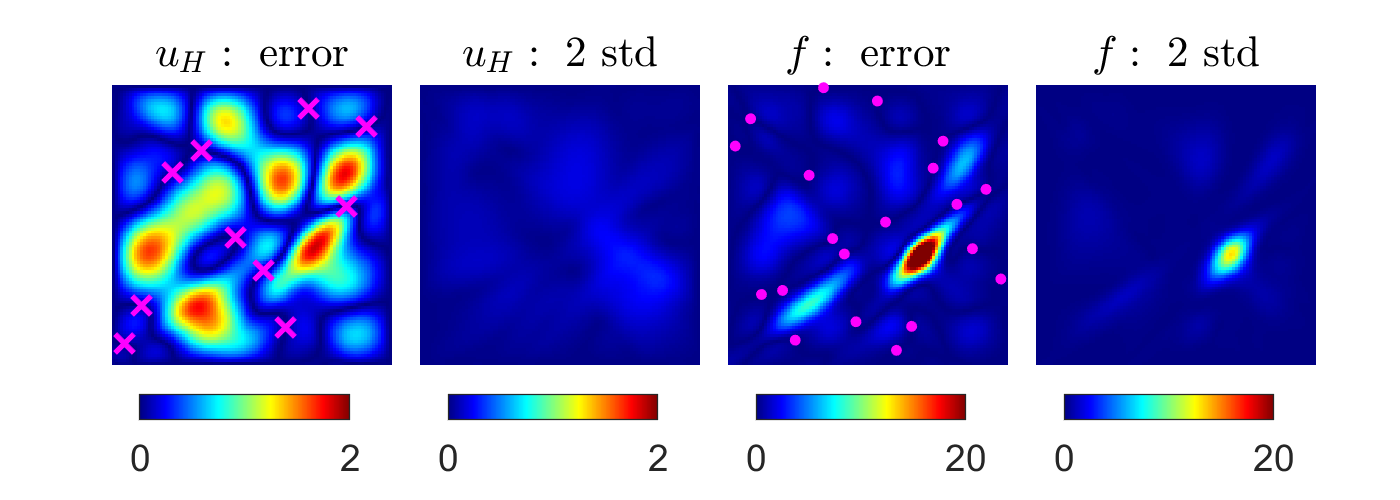}}
    \caption{
    2D inverse problem: 
    (a) Predicted means and uncertainties for $u$ and $f$ using multi-fidelity data: $\epsilon_L \thicksim 0$, $\sigma = 0.7$. 
    (b) Predicted means and uncertainties for $u$ and $f$ using multi-fidelity data: $\epsilon_L \thicksim \mathcal(0, 0.05^2)$, $\sigma = 2.8$.
    (c) Predicted means and uncertainties for $u$ and $f$ using high-fidelity data: $\sigma = 2.3$.
    Magenta cross: high-fidelity training data for $u$, Magenta circle: high-fidelity training data for $f$.
    }
    \label{fig:2d_inverse}
\end{figure}

In these two cases, 6,000 randomly selected  low-fidelity data are employed. In addition, 20 high-fidelity measurements for $u$ at each boundary are used to provide boundary conditions. Furthermore, we assume that 10 and 20 sensors for $u$ and $f$ are available, which are randomly located inside the computational domain. In the DNN for the low-fidelity data we employ 2 hidden layers with 40 neurons per layer, while for the BNN we use 1 hidden layer with 50 neurons. The prior distribution for each hyperparameter is obtained by maximizing the ELBO in the VI, and a standard normal distribution is again utilized as the prior for $k$, which is the same as in Sec. \ref{sec:1Dpde}. As before, we also test the performance of the single-fidelity modeling, i.e., we employ the same BNN using the high-fidelity data only.

The predicted means and uncertainties for $u$ and $f$ using the multi-fidelity modelings are illustrated in Figs. \ref{fig:2d_inversea}-\ref{fig:2d_inverseb}. As shown, the computational errors for $u$ and $f$ in both cases are mostly bounded by the two standard deviations. Furthermore, we present the results from single-fidelity modeling in Fig. \ref{fig:2d_inversec}. We see that the computational errors are about one order larger than the results from multi-fidelity modelings. Finally, the predicted values  of $k$  are displayed in Table \ref{tab:2d_kpred}; we observe that (1) the computational errors in both multi-fidelity modelings are bounded by the one standard deviation, (2) the computational errors increase with the noise scale in the low-fidelity data, consistent with the predicted uncertainties, and (3) the results from multi-fidelity modeling are more accurate than those from single-fidelity modeling.

\begin{table}[h]
\centering
{\footnotesize
\begin{tabular}{c|cccc}
\hline \hline
    & {MF, $\epsilon_{u_L} = 0$} & {MF, $\epsilon_{u_L} \thicksim \mathcal{N}(0, 0.05^2)$}   & {SF} \\
  \hline
  Mean   & 0.981 & 1.058 & 2.287 \\
   Std   & {\footnotesize $3.02 \times 10^{-2}$} & {\footnotesize $5.43 \times 10^{-2}$}  &   {\footnotesize $1.915 \times 10^{-1}$}\\ 
  \hline \hline
\end{tabular}
}
\caption{\label{tab:2d_kpred}
2D inverse PDE problem: Predicted mean and standard deviation for $k$, where $k = 1$ is the exact solution.  MF, $\epsilon_{u_L} = 0$: multi-fidelity modeling with $\epsilon_{u_L} = 0$; MF, $\epsilon_{u_L} \thicksim \mathcal{N}(0, 0.05^2)$: multi-fidelity modeling with $\epsilon_{u_L} \thicksim \mathcal{N}(0, 0.05^2)$; SF: single-fidelity modeling.
}

\end{table}

\subsection{Active learning for reducing uncertainty}

\subsubsection{Function approximation}
\label{sec:active_func}
The low- and high-fidelity data are generated using the same functions as in Sec. \ref{sec:1dfunc}. In addition, we also consider that we have two types of low-fidelity data, i.e., (1) simulation data with no noise, and (2) measurements with large noise, $\epsilon_L \thicksim \mathcal{N}(0, 0.05^2)$. In both cases, we again employ 100 uniformly distributed data at the low level for training. As for the high-fidelity training samples, the noise is also assumed to be Gaussian, i.e., $\epsilon_H \thicksim \mathcal{N}(0, 0.01^2)$. We assume that we only have 10 random high-fidelity measurements at the beginning. The architectures for the NNs are the same as used in Sec. \ref{sec:1Dpde}, i.e., 2 hidden layers with 20 neurons per layer for the low-fidelity DNN, and 1 hidden layer with 50 neurons for the BNN. Furthermore, the priors for the aforementioned cases are $\sigma = 1.2$ and 1.4, respectively.

\begin{figure}[H]
    \centering
    \subfigure[]{\label{fig:activea}
    \includegraphics[width=0.45\textwidth]{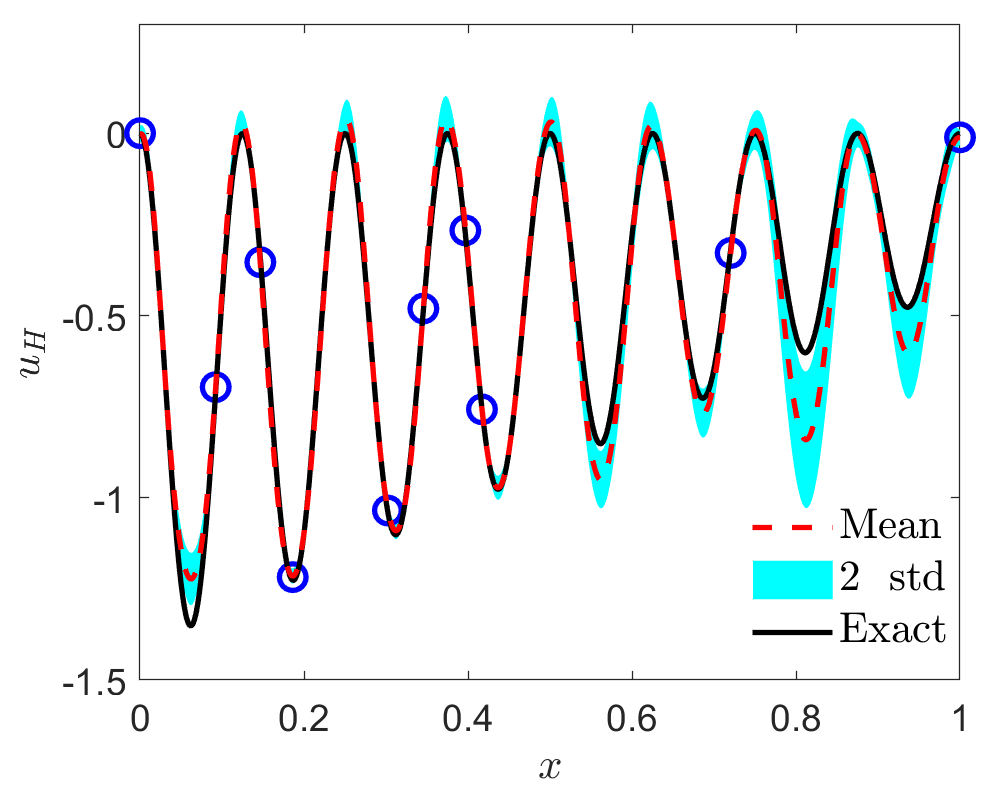}}
     \subfigure[]{\label{fig:activeb}
    \includegraphics[width=0.45\textwidth]{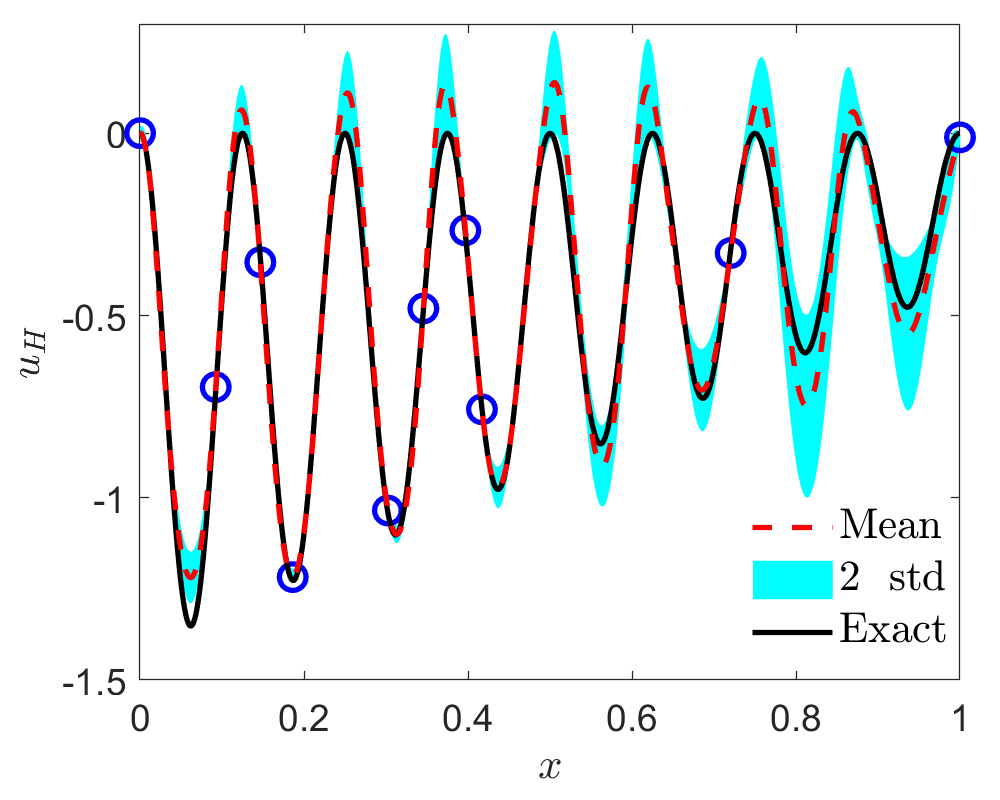}}
    \subfigure[]{\label{fig:activec}
    \includegraphics[width=0.45\textwidth]{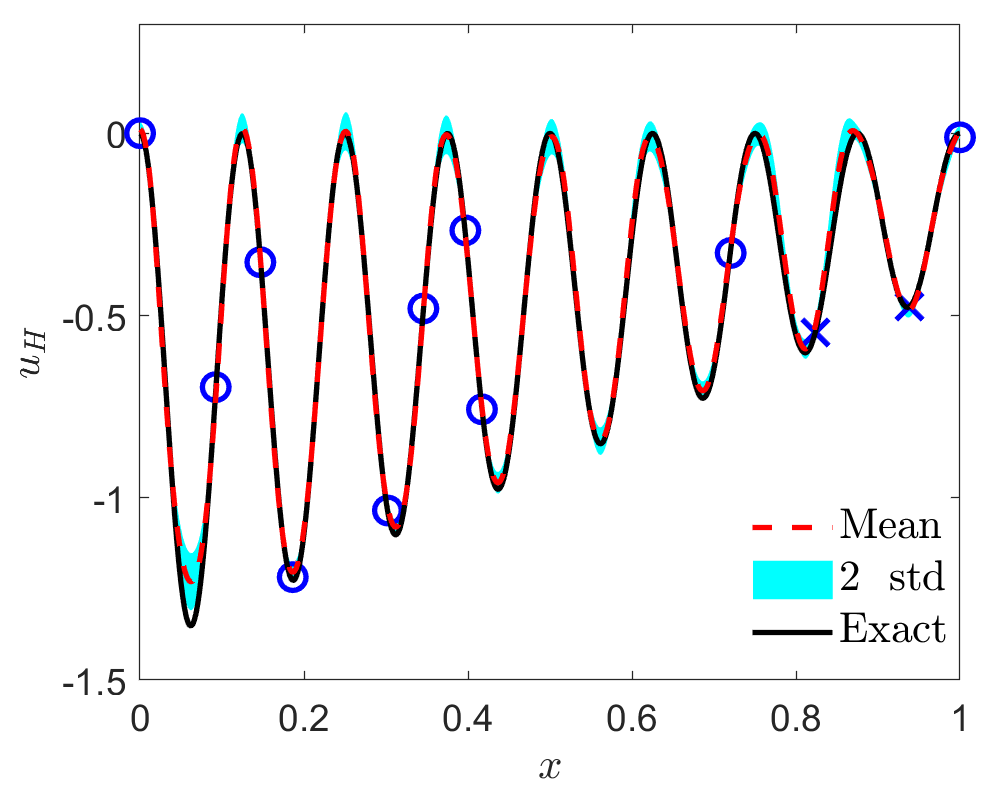}}
     \subfigure[]{\label{fig:actived}
    \includegraphics[width=0.45\textwidth]{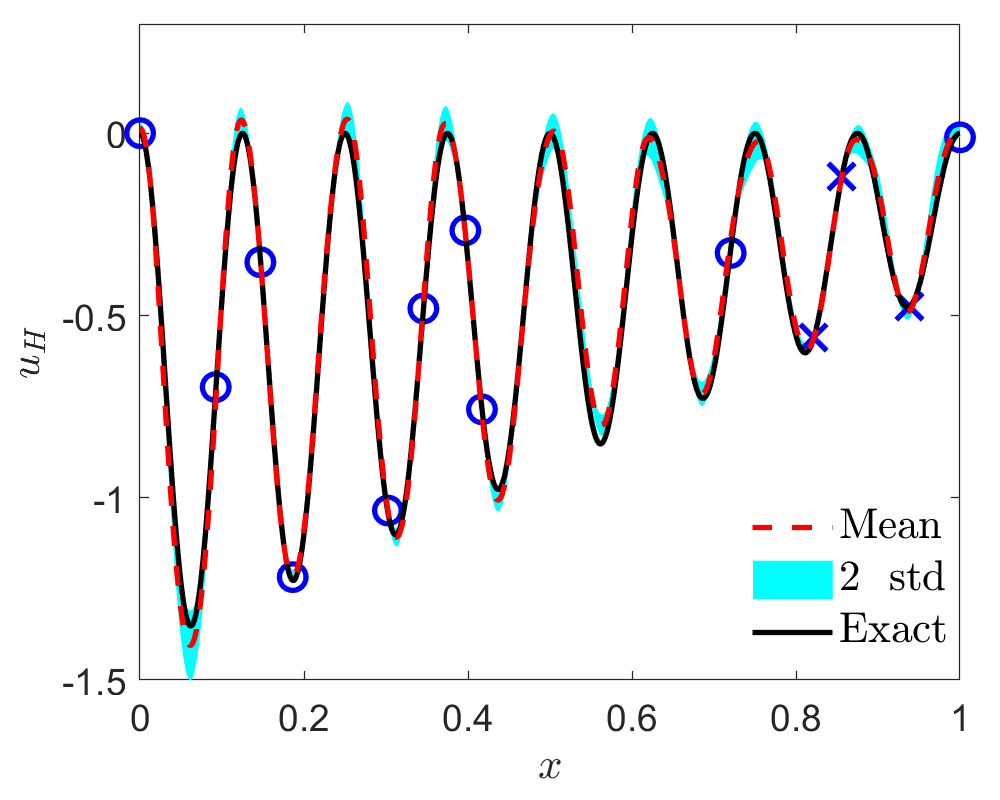}}
    \caption{
    Active learning for function approximation.  
    (a) Predicted means and uncertainties for high-fidelity profiles using 10 initial high-fidelity data with $\epsilon_L \thicksim 0$. 
    (b) Predicted means and uncertainties for high-fidelity profiles using 10 initial high-fidelity data with $\epsilon_L \thicksim \mathcal{N}(0, 0.05^2)$. 
    (c) Predicted means and uncertainties for high-fidelity profiles with 12  high-fidelity data with $\epsilon_L \thicksim 0 $. 
    (d) Predicted means and uncertainties for high-fidelity profiles with 13 high-fidelity data with $\epsilon_L \thicksim \mathcal{N}(0, 0.05^2)$. Blue circle: initial high-fidelity training data, Blue cross: added high-fidelity observations.}
    \label{fig:active}
\end{figure}

The predicted high-fidelity profiles are illustrated in Figs. 
\ref{fig:activea}-\ref{fig:activeb}. As shown, the predicted means agree well with the exact solution for $x < 0.44$. However, the discrepancy becomes larger for $x > 0.44$ since we only have 2 training data in this region; the predicted uncertainty is larger for $x > 0.44$ than in the region of $x < 0.44$. To enhance the predicted accuracy, we can add more noisy high-fidelity training data. In the present study, we employ the same strategy to acquire more high-fidelity data as in \cite{raissi2017inferring}, i.e., adding a new observation $x^*$ for $u_H$ at the location where the posterior variance is maximized. Finally, the iterations of adding high-fidelity measurements terminate as the predicted variance is smaller than $0.05^2$. As shown in Figs. \ref{fig:activec}-\ref{fig:actived}, the predicted high-fidelity profiles are significantly improved after more high-fidelity observations are added in both cases.

\subsubsection{Inverse PDE problem}
\begin{figure}[H]
    \centering
    \subfigure[]{\label{fig:active_pdea}
    \includegraphics[width=0.95\textwidth]{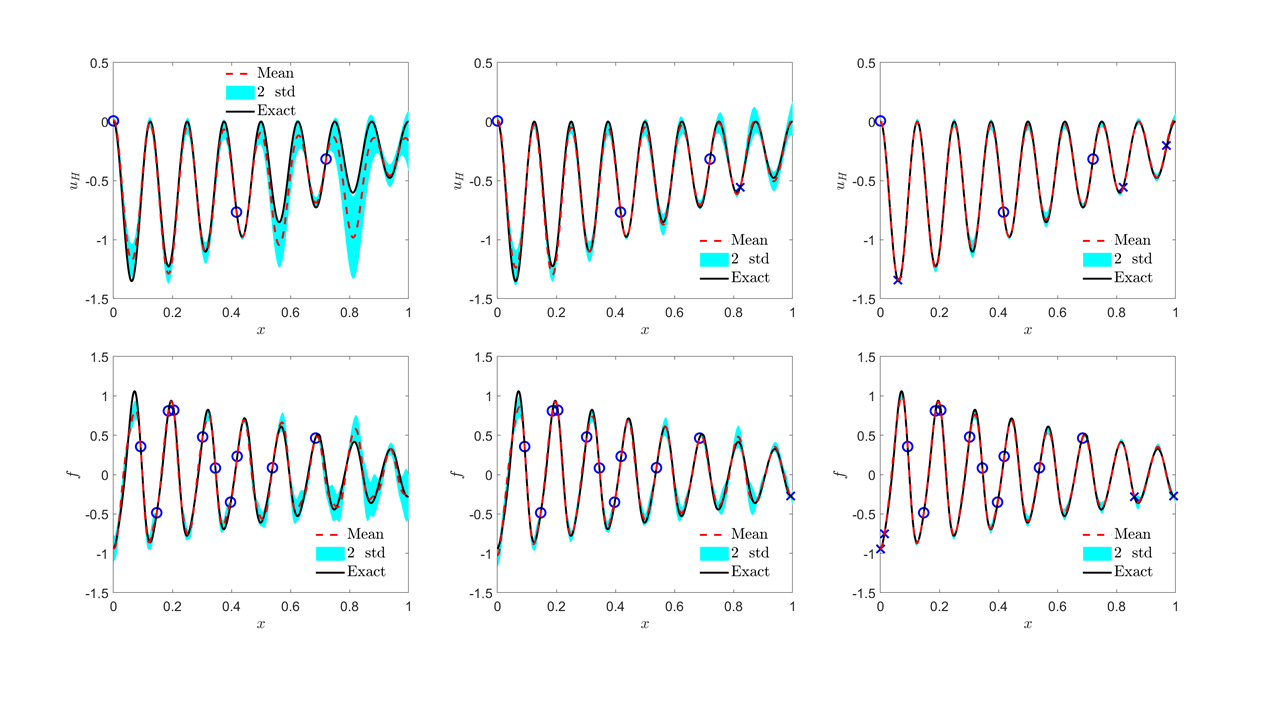}}
    \subfigure[]{\label{fig:active_pdeb}
    \includegraphics[width=0.46\textwidth]{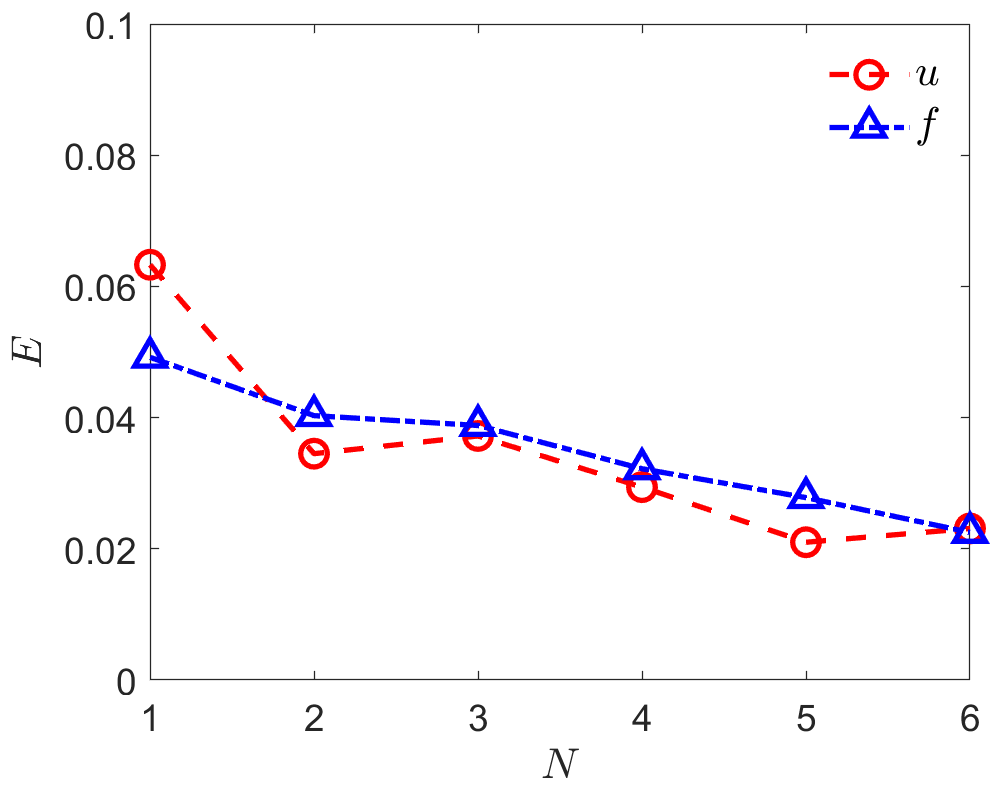}}
    \subfigure[]{\label{fig:active_pdec}
    \includegraphics[width=0.46\textwidth]{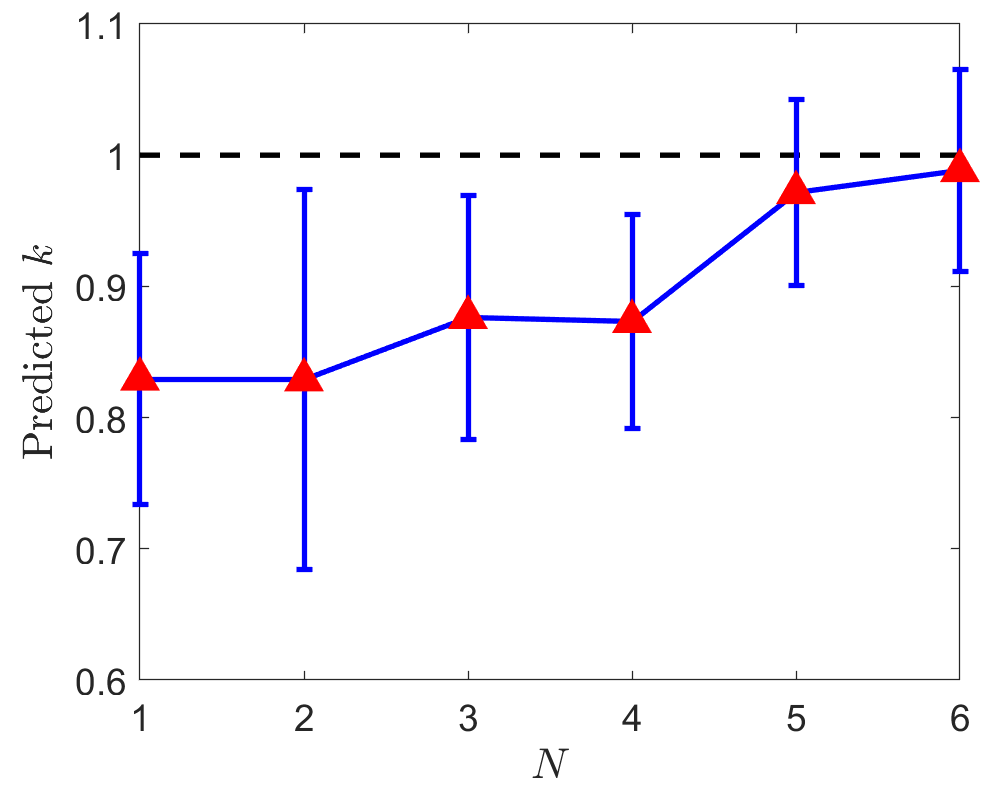}}
    \caption{
    Active learning for 1D inverse diffusion-reaction system.
    (a) Predicted  $u$ and $f$ for $N = 1$, 2, and 5. Blue circle: initial training data; Blue cross: added training points based on the uncertainty.
    (b) Errors for $u$ and $f$ for different iterations.
    (c) Predicted $k$ at different iterations. $N$ represents the number of iterations. Red triangle: predicted mean; error bar: $\pm 2 \sigma$ (i.e., two standard deviations). 
    }
    \label{fig:active_pde}
\end{figure}

Here we employ the same one-dimensional diffusion-reaction system  in Sec. \ref{sec:1Dpde} to demonstrate the concept of the active learning for inverse PDE problem. We assume that we have 3 and 10 high-fidelity measurements for $u$ and $f$ as the initial training set, respectively. In addition, both measurements are randomly distributed in $x \in [0,1]$, as shown in Fig. \ref{fig:active_pdea}. The measurement errors for $u$ and $f$ are also assumed to be Gaussian, i.e., $\epsilon_u \thicksim \mathcal{N}(0, 0.01^2)$ and $\epsilon_f \thicksim \mathcal{N} (0, 0.01^2)$. We also assume that we are able to obtain the low-fidelity solution for $u$ with a guessed $f$, which is the same as the one used in Sec. \ref{sec:1Dpde}, i.e., $u_L(x) = \sin(8 \pi x)$. The low-fidelity data here are assumed to be from simulations, which are noise free. In addition, 100 uniformly distributed samples are employed as the low-fidelity training data. The architectures for the neural networks utilized here are the same as in Sec. \ref{sec:1Dpde}. In addition, the standard deviation for the prior distribution is set as $\sigma = 1.4$.

We start with the initial training set to infer the reaction rate as well as reconstruct the profiles of $u$ and $f$. As shown in Fig. \ref{fig:active_pdea} (first column), neither the predicted $u$ or $f$ are satisfactory, which can be reflected by the large predicted uncertainties. To reduce the uncertainty and enhance the accuracy, we then keep adding  more high-fidelity noisy samples for both $u$ and $f$ using the same strategy as in Sec. \ref{sec:active_func}.  As displayed in Figs. \ref{fig:active_pdea}, the discrepancy between the predicted $u$ and $f$  and the exact solutions become smaller as more training samples are used.

To demonstrate the effect of the training samples on the prediction accuracy quantitatively, we compute the error (i.e., $E$) for $u$ and $f$ as 
\begin{align}
    E = \frac{1}{N}\sqrt{\sum^N_{i=1}|\psi_i - \tilde{\psi}_i|^2}, 
\end{align}
where $N = 1,000$ is the number of predictions, and $\psi$ ($\psi = u/f$) and $\tilde{\psi}$ represent the exact and predicted results, respectively. As observed in Fig. \ref{fig:active_pdeb}, the computational error decreases with the increasing number of iterations. The prediction accuracy for $k$ also improves as the predicted $u$ and $f$ become more accurate as expected (Fig. \ref{fig:active_pdec}).

\section{Summary}
\label{sec:summary}
We presented a multi-fidelity Bayesian neural network (MBNN), which is capable of assimilating a large set of low-fidelity data and scarce high-fidelity data for function approximation as well as for inverse PDE problems and enhancing the prediction accuracy. In particular, MBNN consists of three neural networks: the first one is a deep neural network (DNN) to fit the low-fidelity data, while the second one is a Bayesian neural network (BNN) for capturing the correlation between the low- and high-fidelity data, and the last one is used to encode the physical laws. The first DNN is trained using the maximum a posterior probability (MAP), which is able to handle big data. As for the BNN, we first utilize the mean-field variational inference (VI) to obtain informative prior distributions for the hyperparameters, and then we employ the Hamiltonian Monte Carlo method to sample from the posterior distributions. The low-fidelity data used in the present work can be either simulation data or noisy measurements with low accuracy. In addition, the high-fidelity data are noisy measurements with higher accuracy.

We demonstrated two major advantages of the proposed multi-fidelity BNN: (1) The present approach is capable of capturing both linear and nonlinear correlations between the low- and high-fidelity data adaptively; and (2) it can quantify both aleatoric uncertainty associated with noisy data and epistemic uncertainty  associated with unknown parameters.

The multi-fidelity BNN is first applied for function approximations, including one- and four-dimensional functions, and a two-dimensional sea surface temperature case. The results show that the present method is capable of capturing both linear and nonlinear correlation between the low- and high-fidelity data adaptively, and can provide reasonable uncertainties in predictions. We further employ the multi-fidelity BNN to identify the unknown parameters in both one- and two-dimensional diffusion-reaction systems given a small set of noisy high-fidelity  measurements. The results demonstrated that the present method can estimate the unknown parameter as well as reconstruct the solution with high accuracy. In addition, the multi-fidelity BNN outperforms the single-fidelity modeling in all test cases. We then performed active learning for both a function approximation case as well as an inverse PDE problem, i.e., adding more high-fidelity training data at the location where the uncertainty is maximum. The results indicated that the active learning approach can  reduce the uncertainties efficiently and effectively. Finally, we would like to point out that the present method can be easily extended to big data problems since (1) the mean-field VI used to provide priors can handle big data using the minibatch training, and (2) the stochastic Hamiltonian Monte Carlo \cite{chen2014stochastic,ding2014bayesian,ma2015complete} can be used to replace the Hamiltonian Monte Carlo of the present work to enable the minibatch training for posterior sampling.

\section*{Acknowledgement}
This work was supported by the PhILMS grant DE-SC0019453, OSD/AFOSR MURI grant FA9550-20-1-0358, and the NIH grant U01 HL142518.


\bibliographystyle{unsrt}
\bibliography{refs,HB}

\end{document}